\documentclass{article}

 \usepackage[main, final]{neurips_2026}

\usepackage[utf8]{inputenc} 
\usepackage[T1]{fontenc}    
\usepackage{hyperref}       
\usepackage{url}            
\usepackage{booktabs}       
\usepackage{amsfonts}       
\usepackage{nicefrac}       
\usepackage{microtype}      
\usepackage{xcolor}         

\usepackage{comment}

\usepackage{xcolor, soul}
\usepackage{hyphenat}
\usepackage{multirow}        
\usepackage[table]{xcolor}   
\usepackage{adjustbox}       
\usepackage{graphicx} 
\usepackage{tikz}     
\usetikzlibrary{arrows.meta}
\usepackage{amsmath}
\usepackage{amssymb}

\usepackage{natbib}
\usepackage{xspace}
\def\eg{\emph{e.g.,}\xspace} 
\def\ie{\emph{i.e.,}\xspace} 
\usepackage{enumitem}

\title{
AffineGS: Unlocking Accurate Geometry in 3DGS \\ via  
Spatially Varying Affine Rectification
}

\author{
  Tze Ho Elden Tse$^{1,2}$\thanks{Equal contribution.} \quad
  Jizong Peng$^{1}$\footnotemark[1] \quad
  Richard Chen$^{1}$ \quad
  Angela Yao$^{2}$ \\
  $^{1}$dConstruct Robotics  \quad $^{2}$National University of Singapore \\
  \url{https://eldentse.github.io/AffineGS/}
}

\begin{document}

\maketitle
\begin{abstract}

3D Gaussian Splatting (3DGS) has become the leading approach for photorealistic novel view synthesis, yet its underlying geometric accuracy lags behind its visual quality. Existing depth-guided methods attempt to resolve this by regularizing the optimization with monocular depth priors, typically relying on a global linear alignment. Crucially, we reveal a fundamental flaw in this assumption: modern monocular depth networks systematically exhibit region-dependent, piecewise-linear distortions. This phenomenon invalidates standard global alignment strategies, which inevitably warp scene geometry by enforcing a single scale-and-shift transform. To address this, we introduce a spatially varying affine rectification model that dynamically corrects monocular disparity into consistent metric depth using regionally smooth, edge-aware fields. We further derive a normal-aware regularizer that uses spatial gradients to couple the rectified disparity with 3DGS depth and normals. Extensive evaluations demonstrate that our sensor-free approach prevents structural collapse under extreme view sparsity and extracts highly accurate 3D meshes, effectively bridging the gap toward sensor-supervised geometric fidelity while preserving state-of-the-art photorealism.
\end{abstract}
\section{Introduction}
\label{sec:intro}


3D Gaussian Splatting (3DGS)~\cite{kerbl20233d} has emerged a preferred method for fast and photorealistic rendering of novel views. 
However, while 3DGS excels at capturing high-frequency appearance details, its underlying geometric representation remains inherently unstructured. Compared to implicit neural surfaces~\cite{yu2022monosdf,mildenhall2021nerf} 
or classical Multi-View Stereo, the geometry reconstructed by 3DGS is significantly weaker. As noted in \cite{guedon2024sugar,chung2024depth,turkulainen2025dn}, Gaussian primitives often overfit to the training views, leading to floating artifacts, multi-layered surfaces, and locally inconsistent geometry. This lack of structural integrity is highly problematic for downstream applications requiring reliable 3D meshes, such as autonomous driving, robotic perception, and AR/VR.

To address these geometric shortcomings, one line of work leverages physical sensor depth (\eg DN-Splatter~\cite{turkulainen2025dn}) to directly constrain and align the Gaussian primitives. However, acquiring dense, high-quality sensor depth is hardware-dependent and impractical in many real-world applications. As a more scalable alternative, a highly promising line of research~\cite{chung2024depth,xu2025depthsplat} injects monocular depth priors into the optimization process, leveraging the strong structural cues produced by modern foundation depth models. These methods typically operate on a critical assumption: that predicted monocular disparity can be aligned to metric depth using a global linear transform. While this global alignment strategy successfully provides baseline geometric guidance, our extensive experimentation reveals that this core assumption breaks down in complex real-world environments.

\begin{figure}[t]
  \centering
  \includegraphics[width=\textwidth]{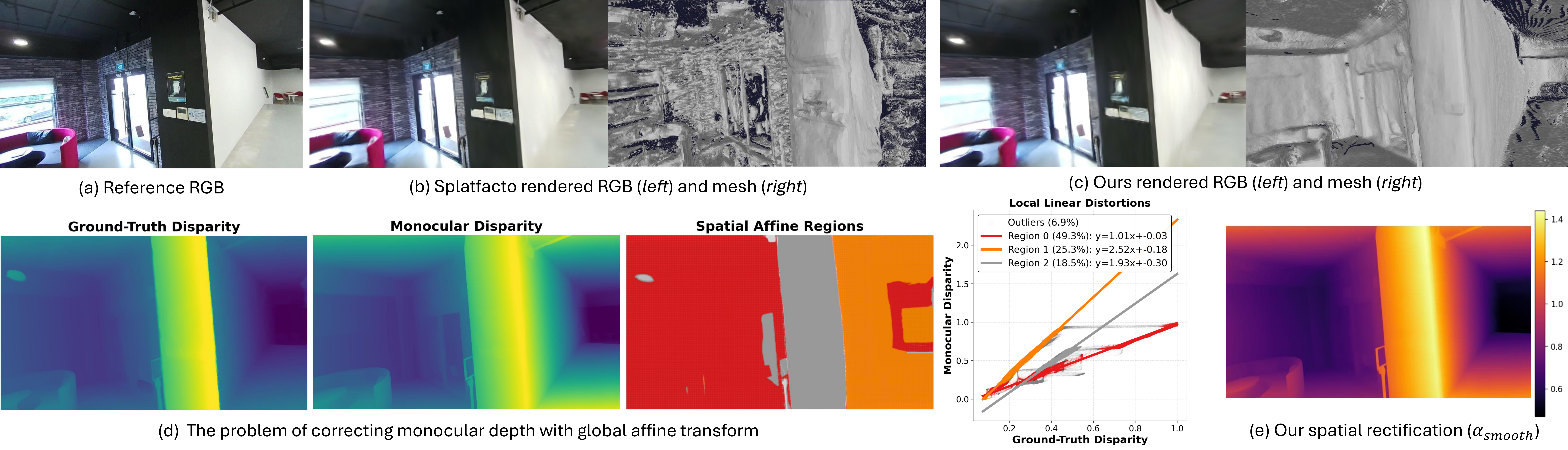}
  \vspace{-1.5em}
  \caption{
  \textbf{Robust 3DGS geometry via spatially varying affine rectification.} 
  (a-c) Standard 3DGS overfits appearance, producing noisy underlying geometry. In contrast, our method extracts clean, highly accurate meshes while maintaining photorealism. (d) Monocular disparity exhibits piecewise-linear distortions, proving that a single global scale-and-shift alignment is fundamentally insufficient (scatter plot). (e) We overcome this by estimating an edge-aware spatial scale field ($\alpha_\text{smooth}$) to rectify the monocular depth prior, yielding geometrically consistent 3D representations.
  }
  \label{fig:teaser}
  \vspace{-1.5em} 
\end{figure}

Specifically, we discover that modern off-the-shelf monocular estimators (\eg Depth-Anything-V2~\cite{yang2024depth}, MoGe~\cite{wang2025moge}, Metric3D-V2~\cite{hu2024metric3dv2}) systematically exhibit region-dependent, piecewise-linear distortions. As they are trained to predict relative depth, their outputs do not map consistently to metric depths. Our findings show that the models tend to fragment scenes into distinct geometric clusters. Fig.~\ref{fig:teaser} shows an example where the predicted disparity fractures into multiple spatial regimes, with breaks along dominant object boundaries or large depth discontinuities. Consequently, attempting to fit a single global scale-and-shift transform will either average the error across these conflicting clusters—resulting in an entirely misaligned scene—or overfit to a single dominant region while severely warping the rest. This phenomenon fundamentally limits the efficacy of global alignment strategies, severely reducing their reliability for geometric supervision in sparse-view settings.

Motivated by this insight, we move beyond global assumptions by introducing a spatially varying affine rectification of monocular disparity. Our key observation is that monocular depth exhibits highly structured distortions: locally smooth, edge-aligned, and piecewise linear. We exploit this structure by dynamically estimating scale-and-shift fields. Crucially, we elegantly recover these spatially varying parameters without relying on any external sensor depth. Specifically, we estimate the local scale factor by comparing the finite-difference spatial gradients of the monocular disparity against the rendered 3DGS depth. We then refine these initial estimates using an edge-aware guided filter, which enforces smoothness within continuous surfaces while preserving sharp discontinuities across object boundaries. This self-contained process rectifies the monocular prior into consistent metric depth prior to geometric supervision. We further integrate this rectified depth with normal-aware constraints derived from projective camera geometry. By formulating our constraints directly on the spatial gradients, we isolate the scale estimation from local shift ambiguities and couple the structural boundaries of the disparity prior with the physical orientations of the 3DGS primitives.

In summary, our contributions are:
\begin{itemize}[nosep]
    \item We show that modern monocular depth predictions suffer from region‑dependent, piecewise‑linear distortions in disparity space which limits global alignment in depth‑guided 3DGS, as a single transform cannot simultaneously regularize distinct geometric clusters.
    \item We present a novel affine rectification model that corrects monocular disparity into consistent metric depth using regionally smooth, edge‑aware fields derived from spatial gradients.
    \item We introduce a gradient‑domain regularizer derived from camera geometry that couples corrected disparity gradients with 3DGS depth and surface normals.
    \item Across diverse experimental settings, our method maintains the strong photometric quality of Gaussian Splatting while substantially improving geometric accuracy, bridging the gap toward sensor-supervised methods without requiring physical depth data.
\end{itemize}

\section{Related Work}

\noindent\textbf{Novel View Synthesis (NVS).} Classical multi-view geometry addressed 3D reconstruction with methods like 
Structure\hyp{}from\hyp{}Motion (SfM) and Multi‑View Stereo (MVS) . 
Tools such as COLMAP~\cite{schonberger2016structure} remain widely used, estimating camera poses and sparse 3D points through epipolar constraints before densifying geometry via stereo matching. While effective, these pipelines depend heavily on feature matching and struggle in low‑texture or non‑Lambertian regions.

With the rise of deep learning, neural reconstruction methods have significantly advanced NVS. 
Implicit scene modeling methods like 
Neural Radiance Fields (NeRF)~\cite{mildenhall2021nerf} represent scenes with a multilayer perceptron and synthesize images via volume rendering. NeRF delivers impressive photorealism and sparked extensive follow‑up research tackling aliasing~\cite{barron2021mip}, efficiency~\cite{reiser2021kilonerf,yu2021plenoctrees,hedman2021baking}, and scalability~\cite{reiser2023merf}. 
Parallel to these developments, explicit or hybrid representations, such as sparse voxel grids~\cite{liu2020neural,sun2022direct}, tensor decompositions~\cite{chen2022tensorf}, and polygonal primitives~\cite{chen2023mobilenerf} have emerged to improve rendering speed by exploiting localized computation.
Among the explicit representations, 3D Gaussian Splatting (3DGS)~\cite{kerbl20233d} stands out for enabling real‑time rendering with high visual fidelity through alpha‑blended rasterization of Gaussian primitives. Its combination of performance, simplicity, and quality has made it a leading representation for novel view synthesis. However, 
3DGS's 
geometric accuracy remains limited and 
geometry improvement is an active 
research direction.

\noindent\textbf{3DGS with Depth Priors.} Recent works 
integrate monocular depth estimation~\cite{ranftl2021vision} into 3DGS to strengthen geometric supervision. 
More advanced monocular estimators, such as Depth-Anything-V2~\cite{yang2024depth}, broaden the range of depth priors available. 3DGS pipelines~\cite{chung2024depth,turkulainen2025dn} typically incorporate monocular depth by jointly optimizing a single global scale-and-shift transform during training. This dynamically aligns the predicted disparity to the rendered depth, allowing the aligned prediction to serve as a geometric regularization target. However, as we demonstrate in Section~\ref{subsec: alpha}, this assumption of global linear alignment does not hold in practice due to the region-dependent, piecewise-linear distortions inherent in monocular depth predictions.


Beyond direct depth supervision, 3DGS extensions leverage geometric cues primarily by modifying the underlying representation. Recent works achieve this by structurally adjusting the primitives into 2D patches or disks~\cite{guedon2024sugar, huang20242d}, appending surface normals as explicit attributes~\cite{gao2024relightable}, merging Gaussian rendering with implicit signed-distance fields~\cite{chen2023neusg}, or introducing joint normal-depth supervision pipelines~\cite{yu2024gaussian, turkulainen2025dn}. In contrast, our method does not modify the Gaussian representation nor rely on rigid global alignment assumptions. Instead, we extract reliable geometric supervision purely by targeting and rectifying the spatially varying distortions inherent to monocular depth priors.

\section{Method}
\label{sec:method}
\definecolor{affblue}{HTML}{1367A7}
\definecolor{afforange}{HTML}{D9771C}
\definecolor{affgreen}{HTML}{2E8B57}
\begin{figure}[t]
\centering
\begin{tikzpicture}[
    x=1cm, y=1cm, font=\footnotesize,
    >={Stealth[length=4pt,width=3.5pt]},
    panel/.style={draw=black!18, fill=black!3, rounded corners=3pt, line width=0.4pt},
    box/.style={draw=affgreen, fill=white, rounded corners=2pt, line width=0.6pt},
    ptitle/.style={font=\footnotesize\bfseries},
    btitle/.style={font=\scriptsize\bfseries, text=affgreen},
    note/.style={font=\scriptsize, text=black!65, align=center},
    tag/.style={fill=white, rounded corners=1pt, inner sep=1.2pt, font=\scriptsize, anchor=north west},
    flow/.style={->, line width=0.7pt},
  ]
  \draw[panel] (0.3,0.8) rectangle (4.4,5.3);
  \draw[panel] (4.6,0.8) rectangle (9.0,5.3);
  \draw[panel] (9.2,0.8) rectangle (13.9,5.3);
  \node[ptitle] at (2.35,5.07) {(a) Rendering and prior};
  \node[ptitle] at (6.8,5.07) {(b) Affine rectification};
  \node[ptitle] at (11.55,5.07) {(c) Geometric supervision};

  \node[font=\scriptsize\bfseries, text=affblue, anchor=west] at (0.45,4.68) {3DGS rendering};
  \draw[affblue, fill=white, rounded corners=2pt, line width=0.6pt] (0.5,3.15) rectangle (1.5,4.15);
  \fill[affblue!40,   rotate around={25:(0.77,3.9)}]  (0.77,3.9)  ellipse (0.2 and 0.075);
  \fill[afforange!45, rotate around={-35:(1.24,3.9)}] (1.24,3.9)  ellipse (0.16 and 0.065);
  \fill[affgreen!40,  rotate around={-8:(1.0,3.64)}]  (1.0,3.64)  ellipse (0.27 and 0.08);
  \fill[afforange!40, rotate around={60:(0.7,3.4)}]   (0.7,3.4)   ellipse (0.15 and 0.065);
  \fill[affblue!45,   rotate around={40:(1.27,3.4)}]  (1.27,3.4)  ellipse (0.18 and 0.07);
  \fill[affblue!30,   rotate around={5:(1.0,3.27)}]   (1.0,3.27)  ellipse (0.12 and 0.04);
  \node[note] at (1.0,2.97) {3D Gaussians};
  \draw[flow, affblue] (1.6,3.72) -- node[note, above] {rasterize} (2.85,3.72);
  \shade[left color=black!60!affblue, right color=affblue!8] (2.95,3.8) rectangle (4.2,4.35);
  \node[tag] at (2.98,4.32) {$\mathcal{D}$};
  \shade[left color=violet!55, right color=cyan!25] (2.95,3.1) rectangle (4.2,3.65);
  \node[tag] at (2.98,3.62) {$n$};

  \draw[black!25, dashed] (0.45,2.75) -- (4.25,2.75);
  \node[font=\scriptsize\bfseries, text=afforange, anchor=west] at (0.45,2.55) {Monocular prior};
  \node[inner sep=0] (rgb) at (1.1,1.65) {\includegraphics[width=1.2cm]{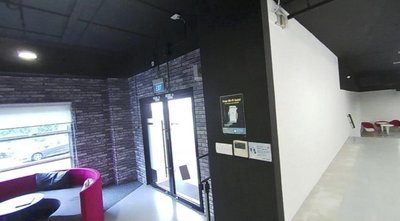}};
  \node[tag] at ([xshift=1pt,yshift=-1pt]rgb.north west) {$I$};
  \node[note] at (1.1,1.1) {training image};
  \node[inner sep=0] (disp) at (3.575,1.65) {\includegraphics[width=1.25cm]{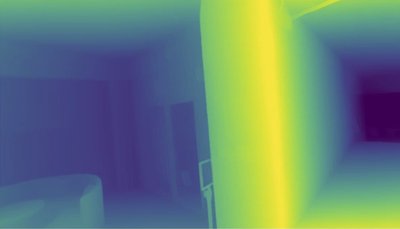}};
  \node[tag] at ([xshift=1pt,yshift=-1pt]disp.north west) {$d$};
  \node[note] at (3.575,1.1) {disparity};
  \draw[flow, afforange] (rgb.east) -- (disp.west);
  \node[note] at (2.32,2.2) {Depth-Anything-V2};

  \draw[flow, affblue] (4.2,4.08) -- node[above, font=\scriptsize, text=affblue] {$\nabla\mathcal{D}$} (4.75,4.08);
  \draw[flow, afforange] (4.2,1.85) -- (4.5,1.85) -- (4.5,3.6) -- (4.75,3.6);
  \node[font=\scriptsize, text=afforange, anchor=east] at (4.47,2.45) {$\nabla d$};
  \draw[flow, afforange, dashed] (4.2,1.45) -- (4.75,1.45);  

  \draw[box] (4.75,3.38) rectangle (8.85,4.85);
  \node[btitle] at (6.8,4.65) {Local affine fields};
  \node at (6.8,4.27) {$\alpha=\|\nabla d\|\,/\,\bigl(d^{2}\,\|\nabla\mathcal{D}\|+\epsilon\bigr)$};
  \node at (6.8,3.91) {$\beta=1/d-\alpha\,\mathcal{D}$};
  \node[note] at (6.8,3.58) {$\alpha$ clamped to its 5th--95th percentiles};
  \draw[flow, affgreen] (6.8,3.38) -- (6.8,3.15);
  \draw[box] (4.75,0.95) rectangle (8.85,3.15);
  \node[btitle] at (6.8,2.95) {Edge-aware guided filter (guide: $d$)};
  \node[inner sep=0] (alph) at (5.6,2.0) {\includegraphics[width=1.5cm]{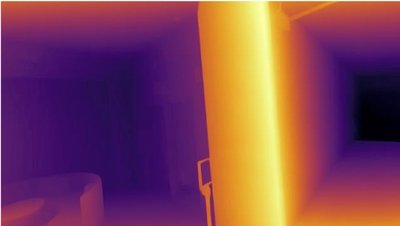}};
  \node[tag] at ([xshift=1pt,yshift=-1pt]alph.north west) {$\alpha_\text{smooth}$};
  \node[note, anchor=west, align=left] at (6.5,2.0) {smooth within\\surfaces, sharp\\at boundaries;\\$\alpha_\text{smooth}$ re-clamped};
  \node[text=affgreen] at (6.8,1.18) {$\alpha_\text{smooth},\ \beta_\text{smooth}$};

  \draw[affgreen, line width=0.7pt] (8.85,2.0) -- (9.1,2.0);
  \draw[affgreen, line width=0.7pt] (9.1,1.55) -- (9.1,4.25);
  \foreach \y in {4.25,2.9,1.55} \draw[flow, affgreen] (9.1,\y) -- (9.35,\y);

  \draw[box] (9.35,3.65) rectangle (13.75,4.85);
  \node[anchor=west] at (9.45,4.63) {\textcolor{affgreen}{$\mathcal{L}_\text{grad}$}\enspace{\scriptsize\color{black!65}disparity-gradient matching}};
  \node at (11.55,4.22) {$\nabla\hat{d}=-\alpha_\text{smooth}\,d^{2}\,\nabla\mathcal{D}$};
  \node[note] at (11.55,3.86) {vs.\ $\nabla d$: magnitude MSE + cosine};

  \draw[box] (9.35,2.3) rectangle (13.75,3.5);
  \node[anchor=west] at (9.45,3.28) {\textcolor{affgreen}{$\mathcal{L}_\text{normal}$}\enspace{\scriptsize\color{black!65}normal-aware gradients}};
  \node at (11.55,2.87) {$\nabla\hat{d}_\text{norm}=-\alpha_\text{smooth}\,d^{2}\,\nabla\mathcal{D}_\text{norm}$};
  \node[note] at (11.55,2.51) {$\nabla\mathcal{D}_\text{norm}$ from $n$ and intrinsics $(f, c)$};

  \draw[box] (9.35,0.95) rectangle (13.75,2.15);
  \node[anchor=west] at (9.45,1.93) {\textcolor{affgreen}{$\mathcal{L}_\text{depth}$}\enspace{\scriptsize\color{black!65}metric re-alignment}};
  \node at (11.55,1.52) {$\hat{\mathcal{D}}=(1-\beta_\text{smooth}\,d)/(\alpha_\text{smooth}\,d)$};
  \node[note] at (11.55,1.16) {$|a\hat{\mathcal{D}}+b-\mathcal{D}|$, $a,b$ by least squares};

  \draw[black!18, fill=white, rounded corners=3pt, line width=0.4pt] (0.3,0) rectangle (13.9,0.6);
  \node at (8.0,0.3) {$\mathcal{L}_\text{total}=\mathcal{L}_\text{photo}+\lambda_1\mathcal{L}_\text{grad}+\lambda_2\mathcal{L}_\text{normal}+\lambda_3\mathcal{L}_\text{depth}$};
  \node[note, anchor=west] at (0.45,0.3) {back-propagate to Gaussians};
  \draw[flow, black!45, dashed] (0.3,0.3) -- (0.12,0.3) -- (0.12,3.65) -- (0.5,3.65);
\end{tikzpicture}
\vspace{-0.6em}
\caption{\textbf{Method overview.} (a) 3DGS renders depth $\mathcal{D}$ and normals $n$, and Depth-Anything-V2 predicts the disparity $d$. (b) The affine fields $\alpha$ and $\beta$ are estimated from disparity and depth gradients and smoothed by a guided filter (Sec.~\ref{subsec: alpha}). (c) They supervise the losses of Secs.~\ref{sec:normal} and~\ref{sec:realignment}. All fields are re-estimated at every iteration.}
\label{fig:method}
\vspace{-1.2em}
\end{figure}

To improve the geometric fidelity of 3D Gaussian Splatting (3DGS), our method leverages monocular depth priors while explicitly accommodating their inherent spatial distortions. Modern monocular estimators 
exhibit region-dependent, piecewise-linear inconsistencies that cannot be resolved via standard global scale-and-shift alignment. Rather than treating these distortions as pure noise, we exploit their 
locally smooth and edge-aligned structure, and introduce a spatially varying affine rectification model (Section~\ref{subsec: alpha}). We then couple these locally consistent metric depth signals with surface-normal and camera-geometry constraints in the gradient domain (Section~\ref{sec:normal}), and anchor the overall scene scale using a direct metric depth re-alignment (Section~\ref{sec:realignment}). Together, these components provide robust, geometry-aware supervision without modifying the underlying 3DGS representation. Fig.~\ref{fig:method} gives an overview of the full pipeline.

\subsection{Spatially Varying Affine Rectification} \label{subsec: alpha}
\noindent\textbf{Motivation.} 
Existing depth-guided 3DGS methods typically assume that monocular disparity can be rectified into metric depth using a single global scale-and-shift transform. However, to quantify the validity of this assumption, we conducted a statistical analysis using the Depth-Anything-V2 prior on the MuSHRoom and ScanNet++ datasets. As shown in Table~\ref{tab:empirical_stats_main}, a single global transform ($K=1$) leaves over 40\% of the scene fundamentally misaligned. Yet, when partitioned into just 3 to 5 distinct local regions, the inlier ratios jump to over 84\% and 91\%, halving the RMSE. This statistically proves that modern monocular estimators exhibit region-dependent, piecewise-linear distortions. Consequently, an uncorrected global prior will inevitably warp the geometry of some objects while aligning others. Rather than treating these distortions as pure noise, we exploit their locally smooth and edge-aligned structure to introduce a spatially varying affine rectification model.
\begin{table}[th]
\centering
\resizebox{0.95\columnwidth}{!}{
\begin{tabular}{l | c c | c c}
\hline
& \multicolumn{2}{c|}{\textbf{MuSHRoom}} & \multicolumn{2}{c}{\textbf{ScanNet++}} \\
\textbf{Alignment Strategy} & \textbf{Inlier Ratio ($\uparrow$)} & \textbf{RMSE ($\downarrow$)} & \textbf{Inlier Ratio ($\uparrow$)} & \textbf{RMSE ($\downarrow$)} \\
\hline
Global Affine ($K=1$) & 58.9\% & 0.074 & 58.6\% & 0.088 \\
Piecewise Local ($K=3$) & 84.9\% & 0.046 & 84.6\% & 0.056 \\
Piecewise Local ($K=5$) & \textbf{92.7\%} & \textbf{0.030} & \textbf{91.8\%} & \textbf{0.039} \\
\hline
\end{tabular}
}
\caption{\textbf{Empirical analysis of disparity distortions.} Evaluated at a 0.02 residual threshold. A single global transform ($K=1$) leaves a significant percentage of the scene misaligned. Partitioning the scene into local linear regions ($K \geq 3$) empirically proves the piecewise-linear nature of monocular priors. See Appendix \ref{sec:appendix_empirical} for a comprehensive evaluation across all thresholds and depth models.}
\label{tab:empirical_stats_main}
\end{table}

\noindent\textbf{Local affine depth modeling.} Let $d(u,v)$ denote the predicted monocular disparity at pixel $(u,v)$, normalized to the min-max range of the current rendering, and let $\mathcal{D}(u,v)$ denote the alpha-blended metric depth rendered by 3DGS. Because monocular networks predict disparity without physical scale, its inverse ($1/d$) acts as a relative depth map. Rather than assuming a strict global alignment, we propose that the relationship between this relative monocular depth and the true metric depth is best modeled locally as a spatially varying affine transformation:
\begin{equation} \label{eq:affine_depth_model}
    \frac{1}{d(u,v)} = \alpha(u,v)\mathcal{D}(u,v) + \beta(u,v),
\end{equation}
where the spatially varying fields $\alpha(u,v)$ and $\beta(u,v)$ correct local scale and shift distortions, respectively. This fundamentally contrasts with existing monocular alignment strategies, which rigidly assume $\alpha$ and $\beta$ are single global constants applied uniformly across the entire image. By differentiating this equation with respect to screen coordinates and applying the chain rule, we obtain:
\begin{equation}
    -\frac{1}{d^2}\nabla d = \alpha \nabla \mathcal{D} + \mathcal{D} \nabla \alpha + \nabla \beta.
\end{equation}
Empirically, the affine correction fields vary more slowly than depth within coherent geometric regions. By applying the local smoothness approximations $\nabla \alpha \approx 0$ and $\nabla \beta \approx 0$, the relationship simplifies to $-\frac{1}{d^2}\nabla d \approx \alpha \nabla \mathcal{D}$, \ie the spatial gradient of the relative monocular depth is locally proportional to the spatial gradient of the rendered metric depth. To robustly estimate the local scale factor $\alpha$ across noisy gradients, we solve for it using the magnitudes of the finite-difference spatial gradients:
\begin{equation}
    \alpha(u,v) = \frac{\|\nabla d\|}{d^2 \|\nabla \mathcal{D}\| + \epsilon},
\end{equation}
where $\epsilon$ is a small constant for numerical stability.


Having solved for the local scale factor $\alpha$ using the spatial gradients, we substitute it back into Eq.~\ref{eq:affine_depth_model} to recover the shift field: $\beta(u,v) = \frac{1}{d(u,v)} - \alpha(u,v)\mathcal{D}(u,v)$. To reject extreme outliers prior to filtering, we strictly clamp $\alpha$ to its $5^\text{th}$ and $95^\text{th}$ percentiles.


\textbf{Edge-aware smoothing via guided filtering.} To ensure regional coherence while respecting sharp object boundaries, we smooth the estimated $\alpha$ and $\beta$ fields using a Guided Filter~\cite{he2012guided}. We use the monocular disparity $d$ as the guide image, which naturally aligns the smoothing process with the scene's structural edges. This promotes local smoothness within coherent depth regions while preventing blurring across object boundaries, effectively capturing the piecewise-linear nature of the monocular distortions. We denote these filtered fields as $\alpha_\text{smooth}$ and $\beta_\text{smooth}$. We use a large filter window (a radius of 550--650 pixels on MuSHRoom; Appendix Sec.~\ref{sec:hyperparams_appendix}), so the filtered fields follow the edges of the guide while their local linear relation to it varies only slowly across the image. Because the guided filter is a local linear fit, it can push the filtered scale field outside the clamping range, including to non-positive values for which the corrected depth in Eq.~\ref{eq:corrected_depth} is undefined. We therefore clamp $\alpha_\text{smooth}$ to the same percentile range after filtering.

\subsection{Normal-Aware Gradient-Domain Regularization} \label{sec:normal}


Depth values are zero-order measurements which are highly susceptible to shift ambiguities. In contrast, surface normals encode first-order geometry. Therefore, we can use these physical surface normals as a strong cue to support our gradient estimation. Specifically, by tying our rectified disparity gradients directly to the expected depth gradients derived from the normals and camera intrinsics, we create a supervision signal that naturally penalizes geometrically implausible local slopes.

\textbf{Coupling disparity and depth gradients.} Using our smoothed affine model, we can analytically predict the target disparity gradient $\nabla \hat{d}$ that perfectly matches the current rendered geometry:
\begin{equation}
    \nabla \hat{d} = -\alpha_\text{smooth} d^2 \nabla \mathcal{D}.
\end{equation}
To enforce consistency between the monocular prior and the rendered geometry, we minimize the discrepancy between the target gradients and the measured disparity gradients. We use a composite loss, applying a Mean Squared Error (MSE) on gradient magnitudes and a cosine similarity to 
enforce angular directional alignment: 
\begin{equation} \label{eq:dispariy_gradient}
    \mathcal{L}_\text{grad} = \left( \|\nabla \hat{d}\| - \|\nabla d\| \right)^2 + \left( 1 - \cos(\nabla \hat{d}, \nabla d) \right).
\end{equation}
For robustness, we clamp both gradient magnitudes to their $5^\text{th}$ and $95^\text{th}$ percentiles before computing the magnitude term.

\textbf{Normal-guided projective constraints.} While $\mathcal{L}_\text{grad}$ transfers structural edges from the monocular prior to the rendered depth map, it only constrains the positions of the Gaussian primitives. Because 3DGS is a volumetric representation, primitives can form a visually smooth alpha-blended depth map while their individual geometric orientations remain chaotic or misaligned with the true surface. This orientation ambiguity severely degrades the quality of extracted 3D meshes. 

To resolve this, we explicitly constrain the first-order orientation of the primitives. We achieve this by anchoring our regularization in projective geometry rather than raw depth; this ensures that spatial depth variations are directly tied to the physical 3D rotations of the Gaussians, preventing internal misalignment. We compute analytic depth gradients directly from the rendered surface normal $n = (n_x, n_y, n_z)$. By enforcing that the local image-space tangents of the back-projected surface are orthogonal to the normal under a pinhole camera with focal lengths $(f_x, f_y)$ and principal point $(c_x, c_y)$, the screen-space derivatives must satisfy:
\begin{align}
    \frac{\partial \mathcal{D}_\text{norm}}{\partial u} &= \frac{-n_x \mathcal{D} / f_x}{n_z + n_x(u-c_x)/f_x + n_y(v-c_y)/f_y}, \\
    \frac{\partial \mathcal{D}_\text{norm}}{\partial v} &= \frac{-n_y \mathcal{D} / f_y}{n_z + n_x(u-c_x)/f_x + n_y(v-c_y)/f_y}.
\end{align}
(The full derivation from a pinhole projection is provided in Appendix Section~\ref{supp:derivation}). 

We substitute these normal-derived gradients into our affine model to generate normal-aware target disparity gradients: $\nabla \hat{d}_\text{norm} = -\alpha_\text{smooth} d^2 \nabla \mathcal{D}_\text{norm}$. By matching these to the monocular disparity gradients, we force the physical orientation of the Gaussians to align with the local slopes of the scene. We apply the same magnitude and cosine similarity loss to this normal branch:
\begin{equation}
    \mathcal{L}_\text{normal} = \left( \|\nabla \hat{d}_\text{norm}\| - \|\nabla d\| \right)^2 + \left( 1 - \cos(\nabla \hat{d}_\text{norm}, \nabla d) \right).
\end{equation}

\subsection{Metric Depth Re-alignment} \label{sec:realignment}

While the gradient-domain losses explicitly supervise local structural fidelity, we additionally provide a direct $\mathcal{L}_1$ depth penalty to prevent the scene from drifting globally. Using the smoothed affine fields, we reconstruct a corrected depth map:
\begin{equation} \label{eq:corrected_depth}
    \hat{\mathcal{D}} = \frac{1 - \beta_\text{smooth} d}{\alpha_\text{smooth} d}.
\end{equation}
Although $\hat{\mathcal{D}}$ captures the correct local geometry, integrating local gradient constraints can leave the reconstruction susceptible to global, low-frequency scale drift relative to the 3DGS coordinate frame. To ensure stable $\mathcal{L}_1$ supervision without inappropriately penalizing arbitrary global scene scales, we perform a global least-squares solve to find optimal scalar values $a$ and $b$ such that $a \hat{\mathcal{D}} + b \approx \mathcal{D}$. This anchors the locally corrected depth to the current global scale of the scene. The final metric depth loss is computed against this globally aligned reconstruction:
\begin{equation}
    \mathcal{L}_\text{depth} = |(a\hat{\mathcal{D}} + b) - \mathcal{D}|.
\end{equation}

The total geometric regularization applied to the 3DGS optimization is the weighted sum of these three components with empirically determined weighting hyperparameters:
\begin{equation}
    \mathcal{L}_\text{geom} = \lambda_1 \mathcal{L}_\text{grad} + \lambda_2 \mathcal{L}_\text{normal} + \lambda_3 \mathcal{L}_\text{depth}.
\end{equation}

\section{Experiments}\label{sec:experiments}

\noindent\textbf{Implementation details.} We train 3DGS using the following loss objective, which is a weighted combination of our proposed geometric regularization:
\begin{equation}
        \mathcal{L}_{\text{total}} = \mathcal{L}_\text{photo} + \lambda_1 \mathcal{L}_\text{grad} + \lambda_2 \mathcal{L}_\text{normal} + \lambda_3 \mathcal{L}_\text{depth} .
    \label{eq:total}
\end{equation}
We empirically set $\lambda_1=0.1$, $\lambda_2=0.08$, $\lambda_3=0.4$ and $\mathcal{L}_\text{photo}$ is the original photometric loss~\cite{kerbl20233d}. On the MuSHRoom scenes \texttt{kokko} and \texttt{sauna}, these default weights produce floaters, so we scale all three weights by 0.5 and 0.25, respectively. On ScanNet++, we use the 0.25 scaling and train at half resolution ($960\times720$). Appendix Sec.~\ref{sec:hyperparams_appendix} lists all per-scene settings. We train all models for 30k iterations. For densification, we use the
unmodified adaptive density control of Splatfacto~\cite{tancik2023nerfstudio}
(gradient-based cloning and splitting, opacity-based pruning, and periodic
opacity reset) with its default settings, identical for our method and all
Gaussian-based baselines. We change no densification logic, so all reported
gains come from our supervision signal alone. To obtain disparity maps, we propagate RGB images through the pre-trained Depth-Anything-V2-Large model~\cite{yang2024depth}. Crucially, the spatially varying affine parameters (\eg the scale factor $\alpha$) used to rectify these monocular priors are iteratively updated alongside the 3DGS scene parameters during optimization. By following~\cite{turkulainen2025dn}, we initialize the Gaussian scene using 1M back-projected points from training dataset sensor depths. The Algorithm column of Tables~\ref{tab:main-table-mesh-mushroom} and~\ref{tab:main-table-mesh-scannetpp} lists the meshing algorithm of each method. We mesh our method, and DN-Splatter (Monodepth) on MuSHRoom, by fusing the depth maps rendered at the training views with Open3D TSDF fusion (voxel size 0.01 and truncation 0.03, in scene units). Methods meshed with Poisson reconstruction extract a total of 2M points from back-projected depth/normal maps and use a depth level of 9; Neusfacto and MonoSDF use marching cubes, and 2DGS uses TSDF fusion. Appendix Sec.~\ref{sec:meshing_appendix} compares Poisson and TSDF meshes of the same trained models. We refer readers to \cite{turkulainen2025dn} for more details.

\noindent\textbf{Baselines.} 
We evaluate across implicit, explicit, and classical representations. Implicit baselines include Nerfacto, Depth-Nerfacto~\cite{tancik2023nerfstudio}, Neusfacto~\cite{yu2022sdfstudio}, and MonoSDF~\cite{yu2022monosdf}. Explicit baselines feature standard Splatfacto~\cite{tancik2023nerfstudio}, geometry-focused SuGaR~\cite{guedon2024sugar} and 2DGS~\cite{huang20242d}, alongside depth-regularized DRGS~\cite{chung2024depth} and DN-Splatter~\cite{turkulainen2025dn}. For a fair sensor-free comparison, we introduce DN-Splatter (Monodepth), which replaces its physical sensor data with dynamically aligned monocular priors (Depth-Anything-V2~\cite{yang2024depth}). Finally, we include classical Poisson surface reconstruction from back-projected sensor depth to contextualize upper-bound mesh performance~\cite{turkulainen2025dn}.

\noindent\textbf{Datasets.} We validate our proposed method across a diverse set of challenging indoor environments. For standard forward-facing captures, we utilize the NeRF-LLFF dataset~\cite{mildenhall2019local}, which comprises distinct real-world scenes. To assess generalization to complex camera movements, we evaluate on the MuSHRoom dataset~\cite{ren2024mushroom}, a real-world indoor benchmark specifically designed with disjoint camera trajectories for training and testing. Finally, to evaluate geometric accuracy, we benchmark on ScanNet++~\cite{yeshwanth2023scannet++}, a large-scale real-world dataset providing dense RGB sequences paired with high-fidelity 3D ground-truth geometry.

\noindent\textbf{Evaluation metrics.} We follow standard practice and report PSNR, SSIM and LPIPS metrics for color images and common depth metrics to analyze depth quality for datasets containing ground truth sensor data. For assessing mesh quality, we follow \cite{turkulainen2025dn} and report accuracy, completion, Chamfer distance, normal consistency and F-scores with a threshold of 5cm.

\subsection{Sparse-View Reconstruction - Table~\ref{tab:views_result}} 

\noindent\textbf{Settings.} Following the evaluation protocol of DRGS~\cite{chung2024depth}, we partition each scene into distinct training and testing sets. For our sparse-view experiments, we optimize the scenes using $k$ randomly sampled training images ($k \in \{2,3,4,5\}$) and evaluate novel-view synthesis on the fixed test set. To ensure statistical reliability, all reported metrics are averaged over ten random seed selections. Additionally, we reference the \textit{Oracle} performance established by DRGS. This oracle serves as an upper bound to demonstrate the theoretical impact of perfect depth geometry; it is generated by optimizing a model on the entire dense image collection (both train and test) to create pseudo-ground-truth depth maps, which are then used in place of standard depth estimates.

\begin{figure}[h]
  \centering
  \includegraphics[width=\textwidth]{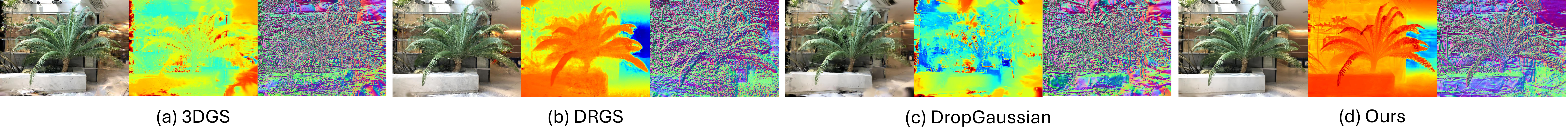}
  \vspace{-1.5em}
  \caption{
  Qualitative comparisons on NeRF-LLFF dataset (more in Appendix Sec.~\ref{sec:llff_appendix}). 
  }
  \label{fig:llff}
  \vspace{-1.5em} 
\end{figure}

\noindent\textbf{Discussions.} While novel view synthesis inherently degrades with view sparsity, our method demonstrates robustness and outperforms all baselines. Notably, DropGaussian's~\cite{park2025dropgaussian} performance severely collapses in the 2-view setting. While their originally reported metrics benefit from a dense, full-scene SfM initialization, our strict evaluation generates SfM points \textit{only} from the available training views, causing their initialization-heavy strategy to fail. Conversely, our approach does not rely on a dense initial point cloud; instead, our spatial distortion correction continuously regularizes the geometry throughout optimization. The feed-forward DepthSplat~\cite{xu2025depthsplat}, evaluated zero-shot, is competitive only on Fern,
the scene closest to its training distribution, and trails our method by 4.3
to 5.0\,dB in mean PSNR at every view count. Crucially, by rectifying these monocular priors, our method unlocks their true potential, bridging the gap to—and occasionally surpassing—the \textit{Oracle} (which relies on dense, multi-view pseudo-GT depth). This occurs because our structural depth priors actively regularize textureless or ambiguous regions where unconstrained multi-view optimization still struggles. Qualitatively (Fig.~\ref{fig:llff}), our method demonstrates a clear reduction in floating artifacts and structural collapse compared to the baselines.

\definecolor{mygreen}{rgb}{0.1, 0.6, 0.1}
\definecolor{myorange}{rgb}{0.8, 0.3, 0.0}

\newcommand{\fst}[1]{\textbf{\textcolor{red}{#1}}}
\newcommand{\snd}[1]{\textcolor{blue}{#1}}
\newcommand{\gain}[1]{\textsuperscript{\textcolor{mygreen}{#1}}}
\newcommand{\loss}[1]{\textsuperscript{\textcolor{myorange}{#1}}}
\newcommand{\blocksep}{\cmidrule(l{0pt}r{0pt}){2-15}}

\begin{table*}[!t]
\centering
\begin{adjustbox}{max width=1.0\textwidth}
\setlength{\tabcolsep}{4pt}
\renewcommand{\arraystretch}{1.05}

\begin{tabular}{c|c|l|llll|llll|llll}
\toprule
\multirow[b]{2}{*}{}
& \multirow[b]{2}{*}{\textbf{Scene}}
& \multirow[b]{2}{*}{\textbf{Methods}}
& \multicolumn{4}{c|}{\textbf{PSNR} $\uparrow$}
& \multicolumn{4}{c|}{\textbf{SSIM} $\uparrow$}
& \multicolumn{4}{c}{\textbf{LPIPS} $\downarrow$} \\
\cmidrule(l{0pt}r{0pt}){4-7}
\cmidrule(l{0pt}r{0pt}){8-11}
\cmidrule(l{0pt}r{0pt}){12-15}
& &
& \textbf{2-view} & \textbf{3-view} & \textbf{4-view} & \textbf{5-view}
& \textbf{2-view} & \textbf{3-view} & \textbf{4-view} & \textbf{5-view}
& \textbf{2-view} & \textbf{3-view} & \textbf{4-view} & \textbf{5-view} \\
\midrule

\multirow{25}{*}{\rotatebox[origin=c]{90}{\textbf{NeRF-LLFF}}}

& \multirow{5}{*}{Fern}
& 3DGS
& 13.03 & 14.29 & 16.73 & 18.59
& 0.336 & 0.408 & 0.517 & 0.603
& 0.476 & 0.389 & 0.296 & 0.217 \\
&
& DropGau.
& 14.71 & 17.58 & 18.04 & 19.27
& 0.413 & 0.565 & \snd{0.620} & \snd{0.670}
& 0.433 & 0.268 & 0.237 & 0.202 \\
& 
& DepthSplat
& \snd{17.77} & 18.86 & 19.20 & 19.71 
& 0.454 & 0.489 & 0.538 & 0.558 
& 0.336 & 0.316 & 0.271 & 0.283 \\
&
& DRGS
& 17.59 & \snd{19.13} & \snd{19.91} & \snd{20.55}
& \snd{0.516} & \snd{0.588} & 0.616 & 0.642
& \snd{0.286} & \snd{0.232} & \snd{0.203} & \snd{0.167} \\
& 
& Ours
& \fst{18.94}\gain{+1.17} & \fst{20.18}\gain{+1.05} & \fst{21.30}\gain{+1.39} & \fst{22.25}\gain{+1.70}
& \fst{0.543}\gain{+0.027} & \fst{0.650}\gain{+0.062} & \fst{0.694}\gain{+0.074} & \fst{0.739}\gain{+0.069}
& \fst{0.266}\gain{-0.020} & \fst{0.189}\gain{-0.043} & \fst{0.154}\gain{-0.049} & \fst{0.121}\gain{-0.046} \\
&
& Oracle
& 18.18 & 20.30 & 20.78 & 21.81 
& 0.524 & 0.636 & 0.654 & 0.701 
& 0.278 & 0.201 & 0.185 & 0.157 \\

\blocksep

& \multirow{5}{*}{Flower}
& 3DGS
& 14.90 & 17.75 & 19.71 & \snd{21.39}
& 0.351 & 0.508 & 0.605 & 0.671
& 0.406 & 0.257 & 0.190 & 0.146 \\
&
& DropGau.
& \snd{16.28} & \snd{19.16} & \snd{19.92} & 21.32
& \snd{0.433} & \snd{0.589} & \snd{0.633} & \snd{0.686}
& \snd{0.320} & \snd{0.194} & \snd{0.171} & \snd{0.138} \\
& 
& DepthSplat
& 11.83 & 15.29 & 17.97 & 18.17 
& 0.154 & 0.317 & 0.425 & 0.460 
& 0.623 & 0.432 & 0.332 & 0.327 \\
&
& DRGS
& 15.92 & 17.80 & 19.15 & 20.45
& 0.395 & 0.445 & 0.538 & 0.576
& 0.414 & 0.376 & 0.323 & 0.293 \\
& 
& Ours
& \fst{19.35}\gain{+3.07} & \fst{21.13}\gain{+1.97} & \fst{22.45}\gain{+2.53} & \fst{23.45}\gain{+2.06}
& \fst{0.567}\gain{+0.134} & \fst{0.657}\gain{+0.068} & \fst{0.722}\gain{+0.089} & \fst{0.754}\gain{+0.068}
& \fst{0.188}\gain{-0.132} & \fst{0.141}\gain{-0.053} & \fst{0.110}\gain{-0.061} & \fst{0.090}\gain{-0.048} \\
&
& Oracle
& 19.71 & 22.16 & 23.26 & 24.65 
& 0.570 & 0.673 & 0.714 & 0.760 
& 0.250 & 0.163 & 0.128 & 0.097 \\
\blocksep

& \multirow{5}{*}{Fortress}
& 3DGS
& 13.87 & 15.98 & 19.26 & 19.98
& 0.363 & 0.492 & 0.609 & 0.631
& 0.389 & 0.283 & 0.201 & 0.191 \\
&
& DropGau.
& 10.94 & 13.92 & 17.96 & 17.36
& 0.419 & 0.557 & 0.685 & 0.693
& 0.490 & 0.273 & 0.193 & 0.197 \\
& 
& DepthSplat
& 13.27 & 16.50 & 18.81 & 18.68 
& 0.309 & 0.339 & 0.409 & 0.401 
& 0.425 & 0.408 & 0.311 & 0.341 \\
&
& DRGS
& \snd{19.80} & \snd{21.85} & \snd{23.07} & \snd{23.72}
& \fst{0.567} & \snd{0.655} & \snd{0.724} & \snd{0.740}
& \snd{0.232} & \snd{0.191} & \snd{0.162} & \snd{0.144} \\
& 
& Ours
& \fst{20.52}\gain{+0.72} & \fst{22.52}\gain{+0.67} & \fst{24.63}\gain{+1.56} & \fst{25.18}\gain{+1.46}
& \snd{0.547}\loss{-0.020} & \fst{0.674}\gain{+0.019} & \fst{0.736}\gain{+0.012} & \fst{0.774}\gain{+0.034}
& \fst{0.187}\gain{-0.045} & \fst{0.133}\gain{-0.058} & \fst{0.119}\gain{-0.043} & \fst{0.111}\gain{-0.033} \\
&
& Oracle
& 23.07 & 24.51 & 26.39 & 26.73 
& 0.654 & 0.728 & 0.787 & 0.797 
& 0.159 & 0.130 & 0.100 & 0.093 \\
\blocksep

& \multirow{5}{*}{Horns}
& 3DGS
& 11.43 & 12.48 & 13.76 & 14.75
& 0.264 & 0.339 & 0.433 & 0.498
& 0.531 & 0.464 & 0.395 & 0.350 \\
&
& DropGau.
& 10.09 & 15.46 & 15.40 & 16.53
& 0.280 & \snd{0.517} & \snd{0.551} & \snd{0.611}
& 0.531 & \snd{0.329} & 0.313 & \snd{0.283} \\
& 
& DepthSplat
& 13.06 & 16.08 & 15.91 & 16.71 
& 0.361 & 0.421 & 0.445 & 0.470 
& 0.418 & 0.376 & 0.380 & 0.383 \\
&
& DRGS
& \snd{15.91} & \snd{16.22} & \snd{18.09} & \snd{18.39}
& \snd{0.420} & 0.466 & 0.527 & 0.565
& \snd{0.362} & 0.349 & \snd{0.306} & 0.296 \\
& 
& Ours
& \fst{16.34}\gain{+0.43} & \fst{19.95}\gain{+3.73} & \fst{21.08}\gain{+2.99} & \fst{22.39}\gain{+4.00}
& \fst{0.473}\gain{+0.053} & \fst{0.671}\gain{+0.154} & \fst{0.730}\gain{+0.179} & \fst{0.776}\gain{+0.165}
& \fst{0.300}\gain{-0.062} & \fst{0.191}\gain{-0.138} & \fst{0.152}\gain{-0.154} & \fst{0.123}\gain{-0.160} \\
&
& Oracle
& 18.56 & 20.08 & 20.88 & 22.52 
& 0.568 & 0.644 & 0.668 & 0.725 
& 0.259 & 0.212 & 0.199 & 0.168 \\
\blocksep

& \multirow{5}{*}{Mean}
& 3DGS
& 13.31 & 15.13 & 17.37 & 18.68
& 0.329 & 0.437 & 0.541 & 0.601
& 0.451 & 0.348 & 0.271 & 0.226 \\
&
& DropGau.
& 13.00 & 16.53 & 17.83 & 18.62
& 0.386 & \snd{0.557} & \snd{0.622} & \snd{0.665}
& 0.444 & \snd{0.266} & \snd{0.229} & \snd{0.205} \\
& 
& DepthSplat
& 13.98 & 16.68 & 17.97 & 18.32 
& 0.319 & 0.391 & 0.454 & 0.472 
& 0.450 & 0.383 & 0.324 & 0.334 \\
& 
& DRGS
& \snd{17.31} & \snd{18.75} & \snd{20.06} & \snd{20.78}
& \snd{0.475} & 0.539 & 0.601 & 0.631
& \snd{0.324} & 0.287 & 0.249 & 0.225 \\
& 
& Ours
& \fst{18.79}\gain{+1.48} & \fst{20.95}\gain{+2.20} & \fst{22.37}\gain{+2.31} & \fst{23.32}\gain{+2.54}
& \fst{0.533}\gain{+0.058} & \fst{0.663}\gain{+0.106} & \fst{0.721}\gain{+0.099} & \fst{0.761}\gain{+0.096}
& \fst{0.235}\gain{-0.089} & \fst{0.164}\gain{-0.102} & \fst{0.134}\gain{-0.095} & \fst{0.111}\gain{-0.094} \\
&
& Oracle
& 19.88 & 21.76 & 22.83 & 23.93
& 0.579 & 0.670 & 0.706 & 0.746
& 0.237 & 0.177 & 0.153 & 0.129 \\

\bottomrule
\end{tabular}
\end{adjustbox}
\caption{Quantitative comparisons on NeRF-LLFF dataset. The best performance (excluding the Oracle) is highlighted in \fst{red}, and the second best in \snd{blue}. The superscript denotes the absolute performance delta of our method against the best-performing baseline.}
\label{tab:views_result}
\vspace{-0.3cm}
\end{table*}

\subsection{Novel-View Synthesis and Depth Estimation - Table~\ref{tab:main-table-nvs-mushroom}}

\noindent\textbf{Settings.} Following the protocol of~\cite{turkulainen2025dn}, we report novel-view synthesis and depth metrics across two distinct evaluation sets (\texttt{left} / \texttt{right}). The \texttt{left} split consists of a test set sampled uniformly every 10 frames from the training sequence, whereas the \texttt{right} split evaluates a disjoint camera trajectory with no training overlap. Results are averaged over 6 scenes.

\begin{figure}[t]
  \centering
  \includegraphics[width=\textwidth]{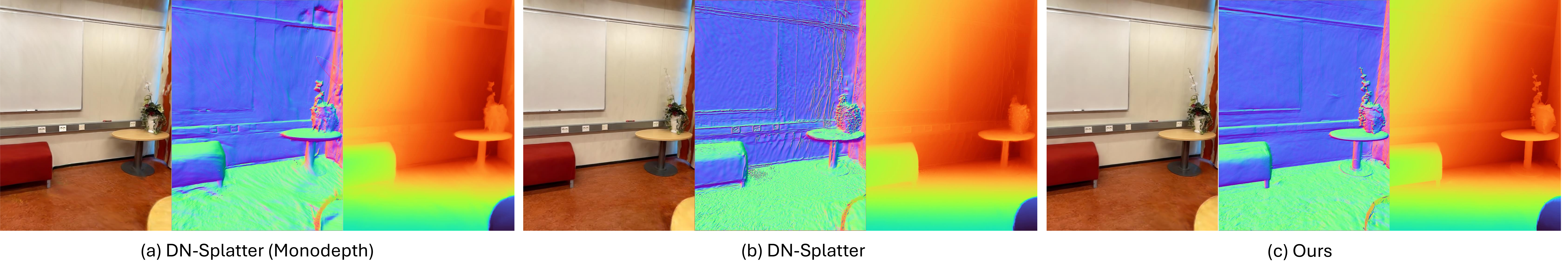}
  \vspace{-1.5em}
  \caption{
  Qualitative comparison on Mushroom dataset (more in Appendix Sec.~\ref{sec:NVS_appendix}). (a) The globally aligned DN-Splatter (Monodepth) baseline suffers from degraded geometries due to structural warping. (c) By rectifying the exact same monocular prior, our method better recovers crisp planar surfaces and sharp complex object boundaries than (b) the sensor-supervised DN-Splatter.
  }
  \label{fig:mushroom_nvs}
  \vspace{-0.5em} 
\end{figure}

\begin{figure}[t]
  \centering
  \includegraphics[width=\textwidth]{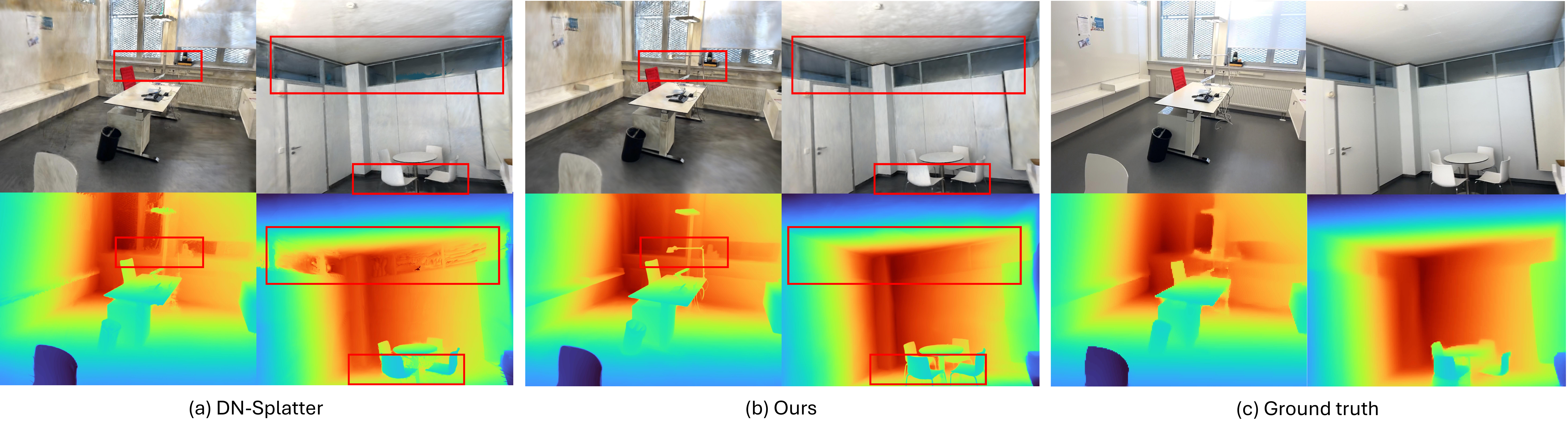}
  \vspace{-1.5em}
  \caption{
  Qualitative comparison on ScanNet++ dataset. We show rendered RGB and depth with red bounding boxes highlighting areas where we generate visually sharper depth, particularly along boundaries of furniture and complex geometry, compared to the DN-Splatter baseline.
  }
  \label{fig:scannet_NVS_appendix}
\end{figure}

\noindent\textbf{Discussions.} Crucially, our proposed method closes the performance gap with methods relying on ground-truth sensor depth, despite strictly operating without it. For novel-view synthesis, we achieve competitive performance, closely matching the sensor-supervised DN-Splatter and outperforming other sensor-free explicit methods like 2DGS and SuGaR. The most significant advantage of our approach lies in its geometric accuracy. While DN-Splatter achieves strong geometry using physical sensor data, its performance severely degrades when forced to rely on monocular priors; the DN-Splatter (Monodepth) baseline experiences a massive structural collapse, yielding an Abs Rel error of 0.3542 / 0.3669 (see Appendix Table~\ref{tab:full_dnsplatter_appendix} for per-scene results). In contrast, we successfully leverage those exact same monocular priors to achieve an exceptional Abs Rel of 0.0551 / 0.0490. This proves that our method can extract high-fidelity geometric constraints from monocular networks without requiring the global sensor alignment that prior works depend on. Finally, Figs.~\ref{fig:mushroom_nvs} and ~\ref{fig:scannet_NVS_appendix} qualitatively corroborate these findings, demonstrating that our geometric regularization preserves crisp novel-view synthesis and prevents structural collapse across both the MuSHRoom and ScanNet++ datasets.

\begin{table*}[t!]
\setlength{\tabcolsep}{4pt}
  \footnotesize%
  \renewcommand*{\arraystretch}{1.05}
  \adjustbox{width=1.0\textwidth}{
\begin{tabular}{clcccc|cccc}
\toprule
 & & Sensor Depth Loss & PSNR $\uparrow$   & SSIM $\uparrow$    & LPIPS $\downarrow$ & Abs Rel $\downarrow$ & Sq Rel $\downarrow$ & RMSE $\downarrow$  & $\delta<1.25$ $\uparrow$ \\
\midrule
\multirow{2}{*}{\rotatebox[origin=c]{90}{NeRF}} & Nerfacto~\cite{tancik2023nerfstudio} & $\textcolor{red}{-}$ & 20.86 / 20.66 & .7859 / .7633 & .2321 / .2702 & .0862 / .0747 & .0293 / .0141 & .0794 / .0667 & .9335 / .9428 \\
& Depth-Nerfacto~\cite{tancik2023nerfstudio} & \textcolor{teal}{$\checkmark$} & 21.24 / 18.93 & .7832 / .7023 & .2414 / .3978 & .0727 / .0563 & .0155 / .0283 & {.0583} / .2840 & .9469 / .9389 \\
\midrule
\multirow{1}{*}{\rotatebox[origin=c]{90}{\hspace{1ex}\mbox{SDF}\hspace{-1.0ex}}} & \rule{0pt}{3ex}MonoSDF~\cite{yu2022monosdf} & \textcolor{teal}{$\checkmark$} & 20.68 / 20.16 & .7357 / .7653 & .3590 / .2261 & .0555 / .0911 & .0263 / .0804 & .2493 / .4535 & .9353 / .8825 \\
\midrule
\multirow{6}{*}{\rotatebox[origin=c]{90}{Explicit}} 
 & DN-Splatter~\cite{turkulainen2025dn} & \textcolor{teal}{$\checkmark$} & 24.58 / 21.89 & {.8558} / {.7984} & .1293 / {.1797} & {.0228} / {.0354} & {.0089} / .0214 & {.1280} / .2032 & {.9854} / .9683 \\
\cline{2-10}
& SuGaR\cite{guedon2024sugar} & $\textcolor{red}{-}$ & 20.52 / 18.18 & .7740 / .7125 & .2427 / .2959 & .1213 / .1174  & .1258 / .1474 & .5562 / .5820 & .8412 / .8564 \\
& 2DGS~\cite{huang20242d} & $\textcolor{red}{-}$ & 22.52 / 21.73 & .8185 / .7898 & .1773 / .1911 & .0864 / .0923 & .0612 / .0583 & .3799 / .3132 & .8820 / .8927 \\
& Splatfacto~\cite{tancik2023nerfstudio} & $\textcolor{red}{-}$ & 24.44 / 21.33 & .8486 / .7821 & .1387 / .2240 & .0787 / .0817 & .0364 / .0521 & .2407 / .2941 & .9072 / .9061 \\
 & DN-Splatter (Monodepth)~\cite{turkulainen2025dn} & $\textcolor{red}{-}$ & 19.26 / 17.24 & .7255 / .6845 & .3338 / .3851 & .3542 / .3669 & .6454 / .7852 & .9351 / 1.103 & .5682 / .5818 \\
\rowcolor{gray!10} & Ours & $\textcolor{red}{-}$ & \textbf{24.49}/ \textbf{22.07} & \textbf{.8570} / \textbf{.7920} & \textbf{.1288} / \textbf{.1733} & \textbf{.0551} / \textbf{.0490} & \textbf{.0176} / \textbf{.0176} & \textbf{.1854} / \textbf{.1953} & \textbf{.9674} / \textbf{.9729} 
\\
\bottomrule
\end{tabular}
}
\vspace{-1em}
\caption{Quantitative comparison on MuSHRoom dataset. We show that our method achieves state-of-the-art geometric accuracy among sensor-free explicit representations, completely avoiding the structural collapse observed in the directly comparable DN-Splatter (Monodepth) baseline. The best results among sensor depth-free methods (indicated by $\textcolor{red}{-}$) are highlighted in \textbf{bold}.
}
\vspace{-1.5em}
\label{tab:main-table-nvs-mushroom}
\end{table*}

\subsection{Mesh Evaluation - Tables~\ref{tab:main-table-mesh-mushroom} and~\ref{tab:main-table-mesh-scannetpp}}

\noindent\textbf{Settings.} We evaluate on MuSHRoom and ScanNet++ datasets. For the MuSHRoom dataset, we report results averaged over 6 scenes: \texttt{coffee\_room}, \texttt{honka}, \texttt{kokko}, \texttt{sauna}, \texttt{computer}, and \texttt{vr\_room}. For the ScanNet++ dataset, the results are averaged over the \texttt{b20a261fdf} and \texttt{8b5caf3398} scenes. Where applicable, we also report our self-reproduced results for DN-Splatter in parentheses.

\begin{figure}[t]
  \centering
  \includegraphics[width=\textwidth]{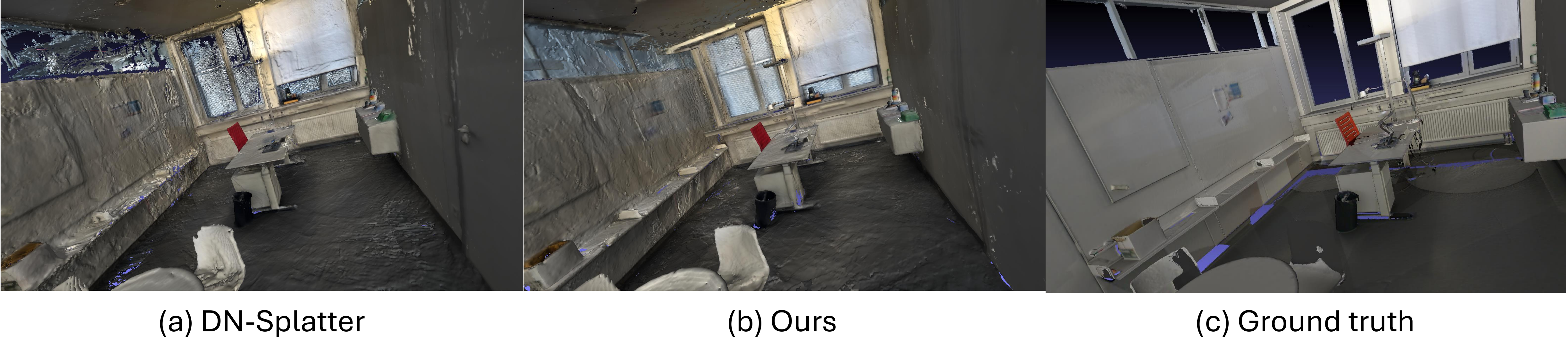}
  \vspace{-1.5em}
  \caption{
  Qualitative mesh comparison on ScanNet++ dataset (more in Appendix Sec.~\ref{sec:mesh_appendix}). Interestingly, we are able to recover window geometries which are absent from ground-truth mesh.
  }
  \label{fig:scannet_mesh}
\end{figure}

\noindent\textbf{Discussions.} The mesh evaluation results validate the structural benefits of our approach. The most critical observation across both datasets is the catastrophic failure of DN-Splatter when deprived of its ground-truth sensor depth. On the MuSHRoom dataset, the DN-Splatter (Monodepth) baseline collapses to an F-score of 0.2765. In contrast, our method maintains geometric accuracies with the same monocular priors (F-score of 0.7097). Furthermore, on the MuSHRoom dataset, our method establishes state-of-the-art performance among all evaluated explicit, sensor-free representations, outperforming dedicated meshing baselines like SuGaR and 2DGS. On ScanNet++, our method again greatly improves over DN-Splatter (Monodepth) and achieves the highest normal consistency among sensor-free methods, while SuGaR and 2DGS obtain lower Chamfer distance and higher F-score. On the \texttt{b20a261fdf} scene, our method, like plain Splatfacto, renders depth about 5\% shorter than the sensor depth, which limits its F-score at the 5\,cm threshold (Appendix Sec.~\ref{sec:limitations}). Across all benchmarks, while sensor-supervised methods naturally dictate the upper bound of performance, our proposed method extracts high-quality geometries without any physical depth data (see Fig.~\ref{fig:scannet_mesh}).

\begin{table*}[th!]
  \setlength{\tabcolsep}{4pt}
  \footnotesize%
  \renewcommand*{\arraystretch}{1.05}
\adjustbox{width=1.0\textwidth}{
\begin{tabular}{clcc|ccccc}
\toprule
 & & Sensor Depth & \multicolumn{1}{c|}{Algorithm} & Accuracy~$\downarrow$ & Completion~$\downarrow$  & Chamfer-$L_1$~$\downarrow$ & Normal Cons.~$\uparrow$  & F-score~$\uparrow$  \\
\midrule
& Traditional Poisson & $\textcolor{red}{-}$ & Poisson & .0399 & .0222 & .0306	& .8688 & .8823 \\
\midrule
\multirow{2}{*}{\rotatebox[origin=c]{90}{NeRF}} 
& Nerfacto~\cite{tancik2023nerfstudio} & $\textcolor{red}{-}$ & Poisson & .0430 & .0578 & .0504 & .7822 & .7212\\
& Depth-Nerfacto~\cite{tancik2023nerfstudio} & \textcolor{teal}{$\checkmark$} & Poisson & .0447 & .0557 & .0502 & .7614 & .6966 \\
\midrule 
\multirow{1}{*}{\rotatebox[origin=c]{90}{\hspace{1ex}\mbox{SDF}\hspace{-1.4ex}}} 
& \rule{0pt}{3ex}MonoSDF~\cite{yu2022monosdf} & \textcolor{teal}{$\checkmark$} & Marching-Cubes & .0310 & .0190 & .0250 & .8846 & .9211 \\
\midrule 
\multirow{6}{*}{\rotatebox[origin=c]{90}{Explicit}} 
& DN-Splatter~\cite{turkulainen2025dn} &  \textcolor{teal}{$\checkmark$} & Poisson &  .0239(.0255) &  .0194(.0291) &  .0216(.0273) &  .8822(.8685) & .9243(.9065) \\
\cline{2-9}
& Splatfacto~\cite{tancik2023nerfstudio} & $\textcolor{red}{-}$ & Poisson & .0749 & .0555 & .0652 & .7727 & .5835 \\
& 2DGS~\cite{huang20242d}& $\textcolor{red}{-}$ & TSDF & .0731 & .0642 & .0687 & .8008 & .6039 \\
& SuGaR~\cite{guedon2024sugar}& $\textcolor{red}{-}$ &  Poisson+IBR & .0656 & .0583 & .0620 & .8031 & .6378 \\
& DN-Splatter (Monodepth)~\cite{turkulainen2025dn} &  $\textcolor{red}{-}$ & TSDF &  .2013 &  .3196 &  .2605 &  .7093 & .2765 \\
\rowcolor{gray!10} & Ours & $\textcolor{red}{-}$ & TSDF & \textbf{.0535} & \textbf{.0437} & \textbf{.0486} & \textbf{.8662} & \textbf{.7097} \\
\bottomrule
\end{tabular}
}
\caption{
Mesh evaluation on MuSHRoom dataset. While DN-Splatter achieves strong geometric accuracy using sensor depth, substituting it with monocular priors (DN-Splatter (Monodepth)) results in severe structural degradation. Our method successfully leverages those same monocular priors without global alignment and outperforms prior explicit baselines like SuGaR and 2DGS. Best results among methods not relying on sensor depth are indicated in \textbf{bold}. Appendix Sec.~\ref{sec:meshing_appendix} compares Poisson and TSDF meshes of the same trained models.
}
\label{tab:main-table-mesh-mushroom}
\vspace{-0.3cm}
\end{table*}

\begin{table*}[th!]
  \setlength{\tabcolsep}{4pt}
  \footnotesize%
  \renewcommand*{\arraystretch}{1.05}
\adjustbox{width=1.0\textwidth}{
\begin{tabular}{clcc|cccccc}
\toprule
 & & Sensor Depth & \multicolumn{1}{c|}{Algorithm} & Accuracy~$\downarrow$ & Completion~$\downarrow$  & Chamfer-$L_1$~$\downarrow$ & Normal Cons.~$\uparrow$  & F-score~$\uparrow$ & Time (min) \\
\midrule
& Traditional Poisson & $\textcolor{red}{-}$ & Poisson & .0593 & .0574 & .0564 & .4410 & .7923 & 2.0\\
\midrule
\multirow{2}{*}{\rotatebox[origin=c]{90}{NeRF}}
& Nerfacto~\cite{tancik2023nerfstudio} & $\textcolor{red}{-}$ &  Poisson & .1305 & .1484 & .1394 & .7153 & \textbf{.4698} & 8.0 \\
& Depth-Nerfacto~\cite{tancik2023nerfstudio} & \textcolor{teal}{$\checkmark$} & Poisson   & .0731 & .1647 & .1189 & .6848 & .5018 & 8.1\\
\midrule 
\multirow{2}{*}{\rotatebox[origin=c]{90}{SDF}}
& Neusfacto~\cite{yu2022sdfstudio} & $\textcolor{red}{-}$ & Marching-Cubes & \textbf{.0736} & .1945 & .1340 & .7159 & .4605 &40.0\\
& MonoSDF~\cite{yu2022monosdf} & \textcolor{teal}{$\checkmark$} & Marching-Cubes & .0303 & .0573 & .0438 & .8881 & .8577 & 47.5\\
\midrule
\multirow{6}{*}{\rotatebox[origin=c]{90}{Explicit}}
& DN-Splatter &  \textcolor{teal}{$\checkmark$} &  Poisson &  .0940(.1226) &  .0395(.0685) & .0667(.0956) & .8316(.7794) &  .7658(.5644) & 36.9\\
\cline{2-10}
& Splatfacto~\cite{tancik2023nerfstudio} & $\textcolor{red}{-}$ & Poisson & .1934 & .1503 & .1719 & .6741 & .1790 &8.9 \\
& 2DGS~\cite{huang20242d} & $\textcolor{red}{-}$ & TSDF & .1272 & \textbf{.0798} & .1035 & .7799 & .4196 & 33.5\\
& SuGaR~\cite{guedon2024sugar} & $\textcolor{red}{-}$ & Poisson + IBR   & .0940 & .1011 & \textbf{.0975} & .7241 & .4367 & 70.0\\
& DN-Splatter (Monodepth) &  $\textcolor{red}{-}$ &  Poisson &  .1798 &  .2087 & .1975 & .7378 &  .1395 & 38.0\\
\rowcolor{gray!10} & Ours & $\textcolor{red}{-}$ & TSDF & .1212 & .0924 & .1068 & \textbf{.8308} & .3207 & 43.4\\
\bottomrule
\end{tabular}
}
\caption{Mesh evaluation on ScanNet++ dataset. While sensor-supervised methods dictate the upper bound performances, our approach achieves the highest normal consistency among sensor-free methods and greatly improves over DN-Splatter (Monodepth), while several baselines achieve better Chamfer distance and F-score.
Best results among sensor-free methods are indicated in \textbf{bold}. 
}
\label{tab:main-table-mesh-scannetpp}
\vspace{-0.1cm}
\end{table*}

\subsection{Ablations}

\noindent\textbf{Proposed geometric regularization (Table~\ref{tab:loss_ablation}).} Unconstrained Splatfacto achieves high PSNR (25.293) but poor geometry (F-score 0.728) due to appearance overfitting. Adding $\mathcal{L}_\text{depth}$ anchors global scale (Abs Rel: $0.059 \to 0.043$), while $\mathcal{L}_\text{normal}$ promotes surface smoothness. Crucially, $\mathcal{L}_\text{grad}$ couples disparity gradients to physical structure, preventing the optimizer from ``cheating'' for photometric gain. Enforcing these joint constraints in our full formulation (All) trades a negligible PSNR drop (24.947 vs.\ 25.293) for state-of-the-art geometric fidelity. Full results and qualitative comparisons are provided in Appendix Table~\ref{tab:full_loss_appendix} and Fig.~\ref{fig:ablation_loss_appendix}.

\begin{table}[h]
    \centering
    \renewcommand{\arraystretch}{1.1}
    \resizebox{0.7\linewidth}{!}{
    \begin{tabular}{l|cc|ccc}
        \toprule
        \multirow{2}{*}{\textbf{Method}} & \multicolumn{2}{c|}{\textbf{Photometry}} & \multicolumn{3}{c}{\textbf{Geometry}} \\
        \cmidrule(lr){2-3} \cmidrule(lr){4-6}
        & PSNR $\uparrow$ & LPIPS $\downarrow$ & Abs Rel $\downarrow$ & Accuracy $\downarrow$ & F-Score $\uparrow$ \\
        \midrule
        Splatfacto (Baseline) & 25.293 & 0.086 & 0.059 & 0.043 & 0.728 \\
        \midrule
        + $\mathcal{L}_\text{grad}$ & 24.322 & 0.123 & 0.085 & 0.054 & 0.673 \\
        + $\mathcal{L}_\text{normal}$ & 25.256 & 0.085 & 0.064 & 0.042 & 0.728 \\
        + $\mathcal{L}_\text{depth}$ & 24.561 & 0.098 & 0.043 & 0.030 & 0.798 \\
        + $\mathcal{L}_\text{grad}$ + $\mathcal{L}_\text{normal}$ & 24.553 & 0.116 & 0.075 & 0.051 & 0.707 \\
        + $\mathcal{L}_\text{grad}$ + $\mathcal{L}_\text{depth}$ & 25.012 & 0.101 & 0.041 & 0.026 & 0.850 \\
        + $\mathcal{L}_\text{normal}$ + $\mathcal{L}_\text{depth}$ & \textbf{25.418} & \textbf{0.084} & \textbf{0.040} & 0.028 & 0.813 \\
        \midrule
        \textbf{All (Ours)} & 24.947 & 0.102 & \textbf{0.040} & \textbf{0.025} & \textbf{0.855} \\
        \bottomrule
    \end{tabular}
    }
    \caption{Ablations on proposed geometric regularization components. 
    Evaluated on the \texttt{coffee\_room} scene from the MuSHRoom dataset. Compared to the Splatfacto baseline, our full formulation (\textit{All}) trades a negligible PSNR drop for geometric accuracies. Best geometric results are \textbf{bolded}.
    }
    \label{tab:loss_ablation}
    \vspace{-0.3cm}
\end{table}

\noindent\textbf{Additional ablations} regarding the sensitivity of monocular depth models, filtering strategies, and gradient matching formulations are provided in Appendix Sections~\ref{sec:mono_prior_ablation_appendix},~\ref{sec:ablation_filter_appendix}, and~\ref{sec:ablation_gradient_formulation_appendix}, respectively.
\section{Conclusion}
In this work, we identified a critical limitation in existing depth-guided 3DGS methods: the reliance on global linear alignment fails to account for the region-dependent, piecewise-linear distortions inherent to modern monocular depth priors. To overcome this, we introduced a novel spatially varying affine rectification framework that dynamically corrects these monocular distortions into geometrically consistent metric depth using edge-aware, locally smooth fields. By coupling this rectified depth with normal-aware gradient constraints, our method explicitly regularizes the physical orientation of 3D primitives without requiring any architectural modifications to 3DGS. Extensive evaluations across the NeRF-LLFF, MuSHRoom, and ScanNet++ datasets demonstrate that our approach significantly outperforms sensor-free depth-regularized baselines, as well as existing explicit baselines on NeRF-LLFF and MuSHRoom. Ultimately, our method bridges the gap toward sensor-supervised geometric fidelity.
\clearpage

\bibliographystyle{plain}
\bibliography{main}
\clearpage


\appendix

\section{Comprehensive Analysis of Monocular Disparity Distortions} \label{sec:appendix_empirical}
In Sec.~\ref{subsec: alpha} of the main paper, we demonstrated that monocular depth priors suffer from region-dependent, piecewise-linear distortions by evaluating RANSAC inlier ratios at a standard residual threshold of 0.02. To rigorously prove that this piecewise behavior is an inherent property of foundation depth models and not merely an artifact of a specific error tolerance, we expand our empirical analysis here. 

\subsection{The Discovery of Piecewise-Linear Disparity Fragmentation}

We uncover a previously uncharacterized, systematic failure mode inherent to modern foundation depth estimators: \textbf{piecewise-linear disparity fragmentation}. While foundation models are highly effective at capturing relative depth ordering, our analysis reveals that their predictions do not map to true metric scale through a unified global transform. Instead, the networks systematically fracture scenes into distinct, spatially coherent affine regimes. 

At first glance, one might assume that allocating more degrees of freedom (increasing the number of local regions, $K$) naturally reduces residual error, akin to fitting a higher-degree polynomial to noisy data. However, it is crucial to distinguish the structured distortions we observe from complex pixel-level noise. If the monocular priors simply suffered from random, unstructured inaccuracies, partitioning the data into $K$ linear models would merely overfit to random, scattered subsets of pixels. 

Our empirical evidence, particularly the spatial coherence masks shown in the main text (Fig.~\ref{fig:teaser}), proves otherwise. The $K$ affine regimes are not random; they are highly structured and correspond directly to coherent physical surfaces. The breaks between these linear regimes align strictly with dominant object boundaries and large depth discontinuities. This proves that foundation models encode relative depth smoothly within continuous physical surfaces, but apply fundamentally different scale-and-shift parameters across different objects.

\subsection{The Mathematical Impossibility of Global Alignment}

\begin{figure}[t]
  \centering
  \includegraphics[width=\textwidth]{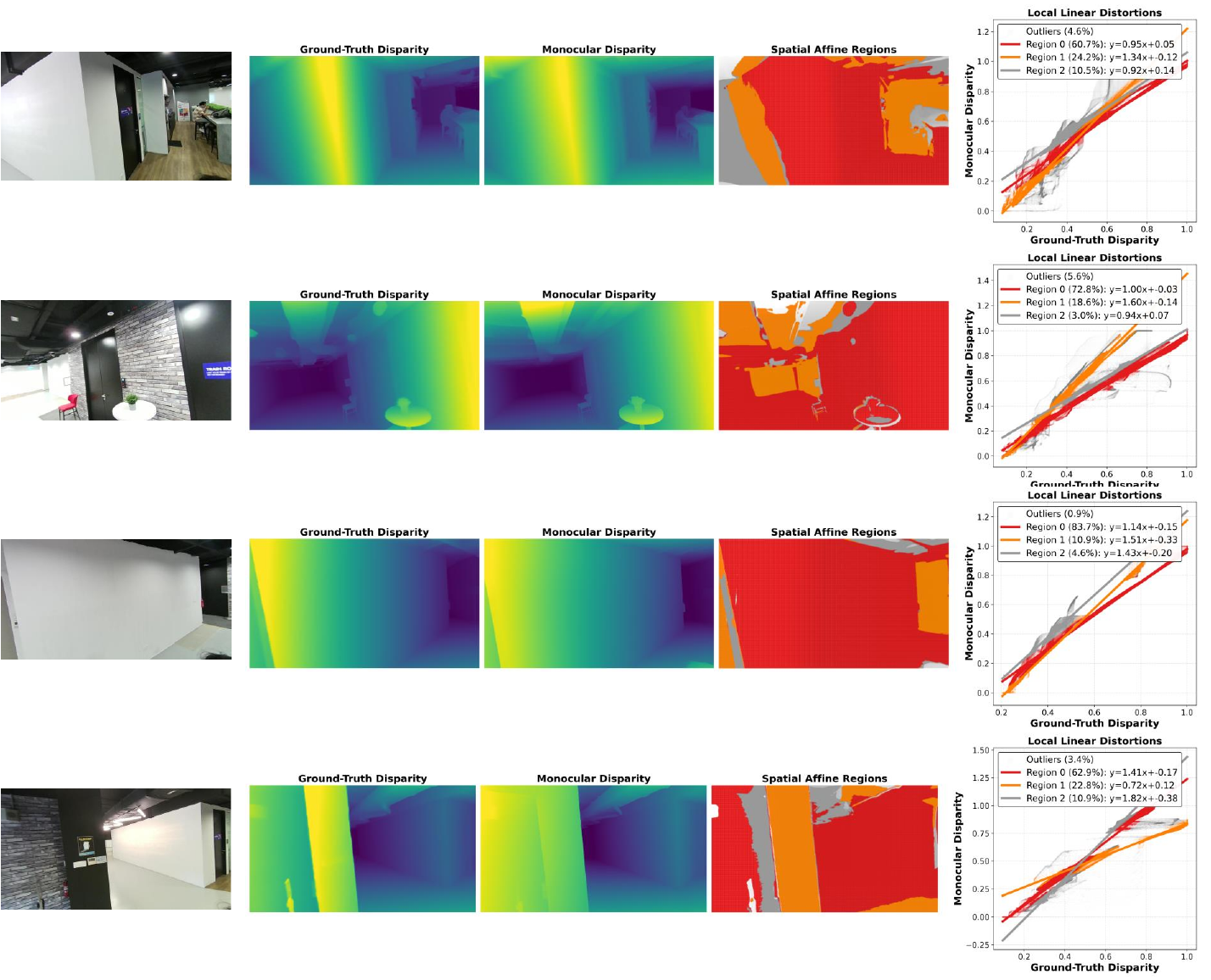}
  \caption{
  Piecewise-linear disparity fragmentation across diverse indoor topologies. Across various scenes, the monocular depth prior consistently fractures into distinct, spatially coherent affine regimes (visualized in red, orange, and grey). The accompanying scatter plots demonstrate that fitting a single global scale-and-shift transform ($K=1$) is mathematically ill-posed, as the optimizer would be forced to average errors across fundamentally incompatible linear clusters.
  }
  \label{fig:affine_stats_4_scenes_appendix}
\end{figure}

This phenomenon fundamentally invalidates the core assumption of existing depth-regularized 3DGS methods. A single global affine transform possesses exactly two degrees of freedom (scale and shift). Consequently, it is mathematically impossible for a single transform to simultaneously reconcile $K$ fundamentally incompatible linear regimes. Global alignment in this context is not merely suboptimal---it is a mathematically ill-posed operation. Attempting to fit a single line through the fragmented disparity space will inevitably force the optimizer to average the error across conflicting spatial clusters, resulting in the severe warping of physical geometry in some regions to satisfy the alignment of others (see Fig.~\ref{fig:affine_stats_4_scenes_appendix}).

\subsection{Empirical Validation Across Foundation Models and Datasets}

\begin{figure}[t]
  \centering
  \includegraphics[width=\textwidth]{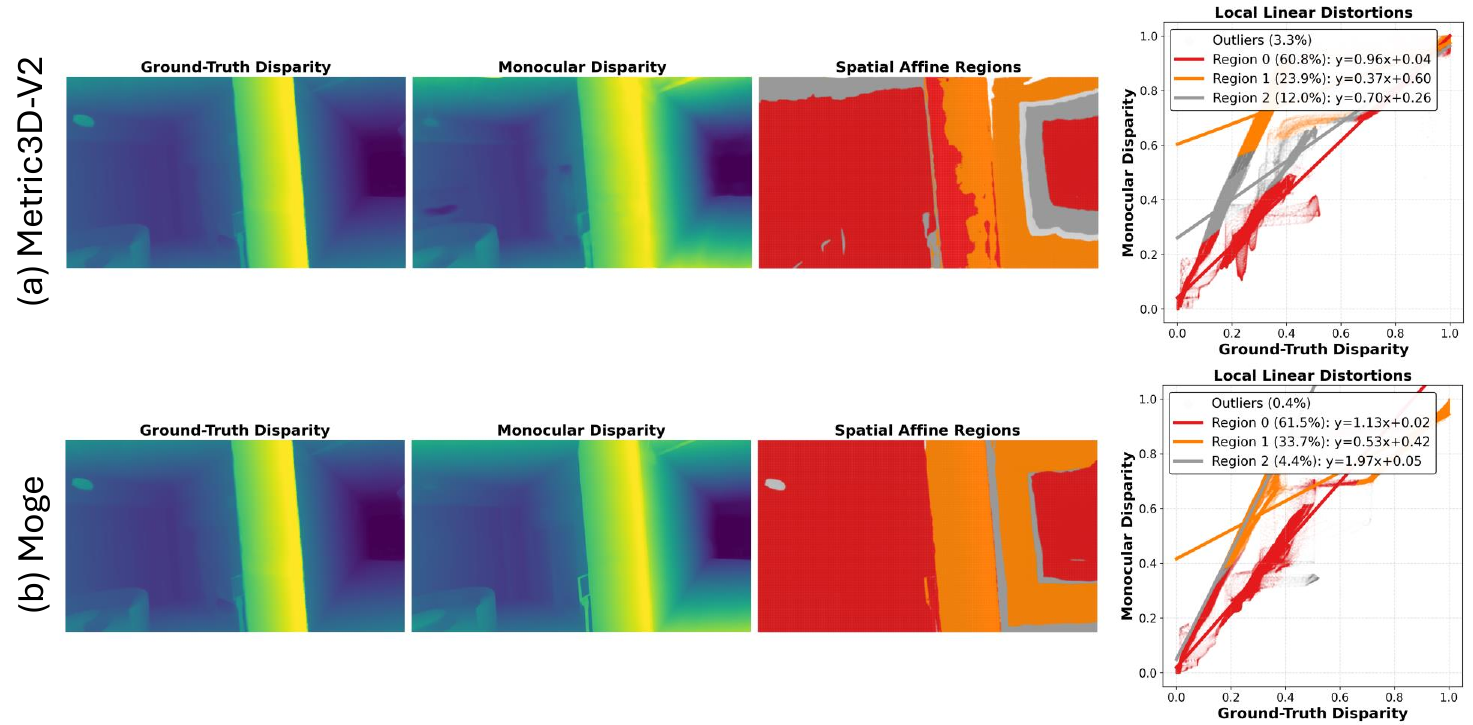}
  \caption{
  We compare the raw disparity predictions of (a) Metric3D-V2 and (b) MoGe against ground-truth sensor depth on the exact same scene. While the distinct network architectures allocate their local scale-and-shift parameters differently, both fundamentally exhibit the same region-dependent, piecewise-linear fragmentation. This confirms that spatial affine distortion is a systemic property of modern foundation depth models. 
  }
  \label{fig:affine_stats_depth_1_appendix}
\end{figure}

To demonstrate that piecewise-linear disparity fragmentation is a systemic characteristic of monocular priors and not an isolated artifact of a specific architecture or dataset, we expand our empirical evaluation across three state-of-the-art foundation models: Metric3D-V2, MoGe, and Depth-Anything-V2. As a direct extension to the visual evidence provided in our main text Fig.~\ref{fig:teaser} (which utilized Depth-Anything-V2), we visually confirm in Fig.~\ref{fig:affine_stats_depth_1_appendix} that both Metric3D-V2 and MoGe exhibit the region-dependent fragmentation behavior.

For each model, we evaluate the affine alignment over the MuSHRoom dataset sequences using iteratively fitted piecewise local linear regions ($K \in \{1, \dots, 8\}$). To rigorously test the spatial alignment, we evaluate the models across three separate residual error thresholds: a strict threshold (0.005), a moderate threshold (0.02), and a forgiving threshold (0.05). Furthermore, to ensure these findings generalize across different capturing hardware and indoor environments, we additionally evaluate Depth-Anything-V2 on the ScanNet++ dataset.

As shown in Tables~\ref{tab:distortions_metric3d} through \ref{tab:empirical_scannet_full}, the global alignment assumption ($K=1$) fails significantly under strict thresholds across all models and datasets. For instance, at a strict tolerance on MuSHRoom, a global transform aligns only 25.6\% of the geometry produced by Depth-Anything-V2. When tested on the entirely different ScanNet++ dataset, it similarly aligns a mere 26.9\% (Table~\ref{tab:empirical_scannet_full}). However, by simply allowing the evaluation to independently align local affine regions ($K \ge 3$), the inlier ratios consistently surge to near-perfect alignment across the board. This drastic improvement confirms that the massive percentage of ``outliers'' rejected by the global transform are not random noise, but rather highly accurate geometric surfaces operating under entirely different affine parameters.

\begin{table*}[h!]
\centering
\resizebox{\textwidth}{!}{
\begin{tabular}{l | c c c | c c c | c c c}
\hline
 & \multicolumn{3}{c|}{\textbf{Strict Threshold (0.005)}} & \multicolumn{3}{c|}{\textbf{Moderate Threshold (0.02)}} & \multicolumn{3}{c}{\textbf{Forgiving Threshold (0.05)}} \\
\textbf{Alignment Strategy} & \textbf{Inlier \% ($\uparrow$)} & \textbf{RMSE ($\downarrow$)} & \textbf{Corr. ($\uparrow$)} & \textbf{Inlier \% ($\uparrow$)} & \textbf{RMSE ($\downarrow$)} & \textbf{Corr. ($\uparrow$)} & \textbf{Inlier \% ($\uparrow$)} & \textbf{RMSE ($\downarrow$)} & \textbf{Corr. ($\uparrow$)} \\
\hline
Global Affine ($K=1$) & 69.4\% & 0.518 & 0.252 & 77.0\% & 0.517 & 0.253 & 92.2\% & 0.517 & 0.254 \\
\hline
Piecewise Local ($K=3$) & 88.7\% & 0.515 & 0.271 & 92.6\% & 0.513 & 0.274 & 98.1\% & 0.373 & 0.452 \\
Piecewise Local ($K=4$) & 92.1\% & 0.502 & 0.289 & 94.8\% & 0.463 & 0.329 & 98.8\% & 0.253 & 0.635 \\
Piecewise Local ($K=5$) & 94.0\% & 0.454 & 0.338 & 96.1\% & 0.394 & 0.414 & 99.1\% & 0.181 & 0.759 \\
Piecewise Local ($K=6$) & 95.3\% & 0.406 & 0.399 & 97.0\% & 0.326 & 0.509 & 99.3\% & 0.137 & 0.839 \\
Piecewise Local ($K=7$) & 96.1\% & 0.356 & 0.468 & 97.6\% & 0.262 & 0.612 & 99.5\% & 0.101 & 0.906 \\
Piecewise Local ($K=8$) & \textbf{96.8\%} & \textbf{0.291} & \textbf{0.564} & \textbf{98.0\%} & \textbf{0.222} & \textbf{0.679} & \textbf{99.6\%} & \textbf{0.086} & \textbf{0.931} \\
\hline
\end{tabular}
}
\caption{\textbf{Comprehensive Statistical Analysis of Distortions (MuSHRoom dataset).} Evaluated across iPhone sequences using the \textbf{Metric3D-V2 (ViT-Large)} prior. The dramatic performance gap between global alignment ($K=1$) and piecewise alignment ($K \ge 3$) persists regardless of the chosen error threshold, proving that foundation models fundamentally exhibit spatially varying scale and shift distortions.}
\label{tab:distortions_metric3d}
\end{table*}

\begin{table*}[h!]
\centering
\resizebox{\textwidth}{!}{
\begin{tabular}{l | c c c | c c c | c c c}
\hline
 & \multicolumn{3}{c|}{\textbf{Strict Threshold (0.005)}} & \multicolumn{3}{c|}{\textbf{Moderate Threshold (0.02)}} & \multicolumn{3}{c}{\textbf{Forgiving Threshold (0.05)}} \\
\textbf{Alignment Strategy} & \textbf{Inlier \% ($\uparrow$)} & \textbf{RMSE ($\downarrow$)} & \textbf{Corr. ($\uparrow$)} & \textbf{Inlier \% ($\uparrow$)} & \textbf{RMSE ($\downarrow$)} & \textbf{Corr. ($\uparrow$)} & \textbf{Inlier \% ($\uparrow$)} & \textbf{RMSE ($\downarrow$)} & \textbf{Corr. ($\uparrow$)} \\
\hline
Global Affine ($K=1$) & 44.1\% & 0.042 & 0.982 & 77.4\% & 0.040 & 0.981 & 92.1\% & 0.039 & 0.982 \\
\hline
Piecewise Local ($K=3$) & 70.5\% & 0.032 & 0.986 & 93.7\% & 0.024 & 0.991 & 98.6\% & 0.020 & 0.994 \\
Piecewise Local ($K=4$) & 76.9\% & 0.029 & 0.988 & 96.0\% & 0.021 & 0.993 & 99.2\% & 0.017 & 0.996 \\
Piecewise Local ($K=5$) & 81.4\% & 0.026 & 0.989 & 97.1\% & 0.018 & 0.995 & 99.5\% & 0.015 & 0.997 \\
Piecewise Local ($K=6$) & 84.7\% & 0.024 & 0.991 & 97.9\% & 0.015 & 0.996 & 99.7\% & 0.013 & 0.998 \\
Piecewise Local ($K=7$) & 87.1\% & 0.023 & 0.991 & 98.4\% & 0.013 & 0.997 & 99.8\% & 0.013 & 0.998 \\
Piecewise Local ($K=8$) & \textbf{89.0\%} & \textbf{0.022} & \textbf{0.992} & \textbf{98.7\%} & \textbf{0.012} & \textbf{0.998} & \textbf{99.9\%} & \textbf{0.012} & \textbf{0.998} \\
\hline
\end{tabular}
}
\caption{\textbf{Comprehensive Statistical Analysis of Distortions (MuSHRoom dataset).} Evaluated across iPhone sequences using the \textbf{MoGe (ViT-Large)} prior. The dramatic performance gap between global alignment ($K=1$) and piecewise alignment ($K \ge 3$) persists regardless of the chosen error threshold, proving that foundation models fundamentally exhibit spatially varying scale and shift distortions.}
\label{tab:distortions_moge}
\end{table*}

\begin{table*}[h!]
\centering
\resizebox{\textwidth}{!}{
\begin{tabular}{l | c c c | c c c | c c c}
\hline
 & \multicolumn{3}{c|}{\textbf{Strict Threshold (0.005)}} & \multicolumn{3}{c|}{\textbf{Moderate Threshold (0.02)}} & \multicolumn{3}{c}{\textbf{Forgiving Threshold (0.05)}} \\
\textbf{Alignment Strategy} & \textbf{Inlier \% ($\uparrow$)} & \textbf{RMSE ($\downarrow$)} & \textbf{Corr. ($\uparrow$)} & \textbf{Inlier \% ($\uparrow$)} & \textbf{RMSE ($\downarrow$)} & \textbf{Corr. ($\uparrow$)} & \textbf{Inlier \% ($\uparrow$)} & \textbf{RMSE ($\downarrow$)} & \textbf{Corr. ($\uparrow$)} \\
\hline
Global Affine ($K=1$) & 25.6\% & 0.086 & 0.967 & 58.9\% & 0.074 & 0.967 & 82.4\% & 0.071 & 0.967 \\
\hline
Piecewise Local ($K=3$) & 50.6\% & 0.060 & 0.974 & 84.9\% & 0.046 & 0.980 & 96.3\% & 0.033 & 0.990 \\
Piecewise Local ($K=4$) & 58.4\% & 0.055 & 0.976 & 89.7\% & 0.037 & 0.985 & 98.0\% & 0.026 & 0.994 \\
Piecewise Local ($K=5$) & 64.4\% & 0.050 & 0.978 & 92.7\% & 0.030 & 0.990 & 98.8\% & 0.022 & 0.996 \\
Piecewise Local ($K=6$) & 69.2\% & 0.047 & 0.980 & 94.6\% & 0.025 & 0.994 & 99.2\% & 0.019 & 0.997 \\
Piecewise Local ($K=7$) & 73.1\% & 0.043 & 0.982 & 95.9\% & 0.022 & 0.995 & 99.5\% & 0.017 & 0.998 \\
Piecewise Local ($K=8$) & \textbf{76.3\%} & \textbf{0.039} & \textbf{0.984} & \textbf{96.8\%} & \textbf{0.019} & \textbf{0.996} & \textbf{99.7\%} & \textbf{0.016} & \textbf{0.998} \\
\hline
\end{tabular}
}
\caption{\textbf{Comprehensive Statistical Analysis of Distortions (MuSHRoom dataset).} Evaluated across iPhone sequences using the \textbf{Depth-Anything-V2 (ViT-Large)} prior. The dramatic performance gap between global alignment ($K=1$) and piecewise alignment ($K \ge 3$) persists regardless of the chosen error threshold, proving that foundation models fundamentally exhibit spatially varying scale and shift distortions.}
\label{tab:distortions_depthanything}
\end{table*}

\begin{table*}[h!]
\centering
\vspace{0.1cm}
\resizebox{\textwidth}{!}{
\begin{tabular}{l | c c c | c c c | c c c}
\hline
& \multicolumn{3}{c|}{\textbf{Strict Threshold (0.005)}} & \multicolumn{3}{c|}{\textbf{Moderate Threshold (0.02)}} & \multicolumn{3}{c}{\textbf{Forgiving Threshold (0.05)}} \\
\textbf{Alignment Strategy} & \textbf{Inlier \% ($\uparrow$)} & \textbf{RMSE ($\downarrow$)} & \textbf{Corr. ($\uparrow$)} & \textbf{Inlier \% ($\uparrow$)} & \textbf{RMSE ($\downarrow$)} & \textbf{Corr. ($\uparrow$)} & \textbf{Inlier \% ($\uparrow$)} & \textbf{RMSE ($\downarrow$)} & \textbf{Corr. ($\uparrow$)} \\
\hline
Global Affine ($K=1$) & 26.9\% & 0.092 & 0.952 & 58.6\% & 0.088 & 0.952 & 81.6\% & 0.084 & 0.952 \\
\hline
Piecewise Local ($K=3$) & 51.8\% & 0.068 & 0.967 & 84.6\% & 0.056 & 0.974 & 95.2\% & 0.040 & 0.985 \\
Piecewise Local ($K=4$) & 59.2\% & 0.065 & 0.969 & 89.1\% & 0.047 & 0.980 & 97.1\% & 0.031 & 0.991 \\
Piecewise Local ($K=5$) & 64.8\% & 0.062 & 0.971 & 91.8\% & 0.039 & 0.985 & 98.1\% & 0.025 & 0.994 \\
Piecewise Local ($K=6$) & 69.3\% & 0.058 & 0.974 & 93.7\% & 0.032 & 0.989 & 98.8\% & 0.022 & 0.995 \\
Piecewise Local ($K=7$) & 73.1\% & 0.053 & 0.978 & 95.0\% & 0.027 & 0.992 & 99.1\% & 0.019 & 0.997 \\
Piecewise Local ($K=8$) & \textbf{76.2\%} & \textbf{0.049} & \textbf{0.981} & \textbf{96.0\%} & \textbf{0.024} & \textbf{0.993} & \textbf{99.4\%} & \textbf{0.017} & \textbf{0.998} \\
\hline
\end{tabular}
}
\caption{\textbf{Comprehensive Statistical Analysis of Distortions (ScanNet++ dataset).} The ScanNet++ benchmark exhibits the exact same piecewise-linear behavior as the MuSHRoom dataset when using the Depth-Anything-V2 (ViT-Large) prior, demonstrating that our findings on monocular distortions generalize across different indoor environments and capturing hardware.}
\label{tab:empirical_scannet_full}
\end{table*}

\section{Derivation of Normal-Aware Depth Gradient Constraints} \label{supp:derivation}

This section provides the full derivation for the normal-aware depth gradients $\frac{\partial \mathcal{D}}{\partial u}$ and $\frac{\partial \mathcal{D}}{\partial v}$ used in Section~\ref{sec:normal} of the main paper. These constraints are derived by enforcing that the local image-space tangents of a back-projected surface patch are strictly orthogonal to the surface normal under a pinhole camera model.

\subsection{Setup and Notation}
Let $(u,v)$ be the continuous screen-space pixel coordinates. We assume a standard pinhole camera with focal lengths $(f_x, f_y)$ and principal point $(c_x, c_y)$. 

A pixel at location $(u,v)$ with metric depth $\mathcal{D}(u,v)$ back-projects into 3D camera coordinates $\mathbf{P}(u,v) = [X, Y, Z]^\top$ according to the pinhole projection equations:
\begin{equation} \label{eq:pinhole}
    X = \frac{u - c_x}{f_x}\mathcal{D}(u,v), \quad Y = \frac{v - c_y}{f_y}\mathcal{D}(u,v), \quad Z = \mathcal{D}(u,v).
\end{equation}

Let $\mathbf{n} = [n_x, n_y, n_z]^\top$ denote the unit surface normal vector at this 3D point. Geometrically, the partial derivatives of the 3D position with respect to the image coordinates, $\frac{\partial \mathbf{P}}{\partial u}$ and $\frac{\partial \mathbf{P}}{\partial v}$, represent the tangent vectors of the local surface patch. By definition, these tangent vectors must be orthogonal to the surface normal $\mathbf{n}$:
\begin{equation} \label{eq:orthogonality}
    \mathbf{n}^\top \frac{\partial \mathbf{P}}{\partial u} = 0, \quad \text{and} \quad \mathbf{n}^\top \frac{\partial \mathbf{P}}{\partial v} = 0.
\end{equation}

\subsection{Horizontal Derivative ($\partial \mathcal{D} / \partial u$)}
To find the horizontal depth gradient, we first differentiate the 3D coordinate vector $\mathbf{P}$ with respect to $u$ (applying the product rule to $X$ and keeping $v$ constant):
\begin{equation} \label{eq:tangent_u}
    \frac{\partial \mathbf{P}}{\partial u} = 
    \begin{bmatrix}
    \frac{\partial X}{\partial u} \\[0.5em]
    \frac{\partial Y}{\partial u} \\[0.5em]
    \frac{\partial Z}{\partial u}
    \end{bmatrix}
    = 
    \begin{bmatrix}
    \frac{\mathcal{D}(u,v)}{f_x} + \frac{u - c_x}{f_x} \frac{\partial \mathcal{D}}{\partial u} \\[0.5em]
    \frac{v - c_y}{f_y} \frac{\partial \mathcal{D}}{\partial u} \\[0.5em]
    \frac{\partial \mathcal{D}}{\partial u}
    \end{bmatrix}.
\end{equation}

Next, we enforce the orthogonality constraint from Eq.~\ref{eq:orthogonality} by taking the dot product with the normal vector $\mathbf{n}$ and setting it to zero:
\begin{equation}
    n_x \left( \frac{\mathcal{D}(u,v)}{f_x} + \frac{u - c_x}{f_x} \frac{\partial \mathcal{D}}{\partial u} \right) + n_y \left( \frac{v - c_y}{f_y} \frac{\partial \mathcal{D}}{\partial u} \right) + n_z \left( \frac{\partial \mathcal{D}}{\partial u} \right) = 0.
\end{equation}

We then distribute $n_x$ and group all terms containing the depth gradient $\frac{\partial \mathcal{D}}{\partial u}$:
\begin{equation}
    n_x \frac{\mathcal{D}(u,v)}{f_x} + \frac{\partial \mathcal{D}}{\partial u} \left( n_x \frac{u - c_x}{f_x} + n_y \frac{v - c_y}{f_y} + n_z \right) = 0.
\end{equation}

Subtracting the isolated term to the right side yields the critical negative sign:
\begin{equation}
    \frac{\partial \mathcal{D}}{\partial u} \left( n_z + n_x \frac{u - c_x}{f_x} + n_y \frac{v - c_y}{f_y} \right) = -n_x \frac{\mathcal{D}(u,v)}{f_x}.
\end{equation}

Finally, dividing by the bracketed term isolates the horizontal gradient:
\begin{equation} \label{eq:final_u}
    \frac{\partial \mathcal{D}}{\partial u} = \frac{-n_x \mathcal{D}(u,v) / f_x}{n_z + n_x \frac{u - c_x}{f_x} + n_y \frac{v - c_y}{f_y}}.
\end{equation}

\subsection{Vertical Derivative ($\partial \mathcal{D} / \partial v$)}
The derivation for the vertical gradient follows identical symmetry. We differentiate $\mathbf{P}$ with respect to $v$ (holding $u$ constant):
\begin{equation} \label{eq:tangent_v}
    \frac{\partial \mathbf{P}}{\partial v} = 
    \begin{bmatrix}
    \frac{u - c_x}{f_x} \frac{\partial \mathcal{D}}{\partial v} \\[0.5em]
    \frac{\mathcal{D}(u,v)}{f_y} + \frac{v - c_y}{f_y} \frac{\partial \mathcal{D}}{\partial v} \\[0.5em]
    \frac{\partial \mathcal{D}}{\partial v}
    \end{bmatrix}.
\end{equation}

Enforcing the orthogonality constraint $\mathbf{n}^\top \frac{\partial \mathbf{P}}{\partial v} = 0$:
\begin{equation}
    n_x \left( \frac{u - c_x}{f_x} \frac{\partial \mathcal{D}}{\partial v} \right) + n_y \left( \frac{\mathcal{D}(u,v)}{f_y} + \frac{v - c_y}{f_y} \frac{\partial \mathcal{D}}{\partial v} \right) + n_z \left( \frac{\partial \mathcal{D}}{\partial v} \right) = 0.
\end{equation}

Grouping the $\frac{\partial \mathcal{D}}{\partial v}$ terms and solving yields the final vertical gradient:
\begin{equation} \label{eq:final_v}
    \frac{\partial \mathcal{D}}{\partial v} = \frac{-n_y \mathcal{D}(u,v) / f_y}{n_z + n_x \frac{u - c_x}{f_x} + n_y \frac{v - c_y}{f_y}}.
\end{equation}

\subsection{Summary}
Equations \ref{eq:final_u} and \ref{eq:final_v} form the analytic depth gradients utilized in the main paper. By substituting these relationships into our affine rectification model, we establish a robust mapping between predicted monocular disparity and the physical geometric constraints dictated by the rendered surface normals.

\section{Implementation Details} \label{sec:hyperparams_appendix}

\subsection{Per-Scene Settings}
Table~\ref{tab:hyperparams_appendix} lists the guided filter parameters and loss weights for every scene. The filter radius is given in pixels at the training resolution, and $\epsilon_\text{gf}$ is the regularization parameter of the guided filter. On MuSHRoom, the default weights ($\lambda_1=0.1$, $\lambda_2=0.08$, $\lambda_3=0.4$) produce floaters on \texttt{kokko} and \texttt{sauna}, so we scale all three weights by 0.5 and 0.25 on these scenes, respectively. Before training, each disparity map is clipped to its $1^\text{st}$ and $99^\text{th}$ percentiles and min-max normalized to $[0, 1]$.

\begin{table}[h]
\centering
\small
\begin{tabular}{llccccc}
\toprule
Dataset & Scene & Radius (px) & $\epsilon_\text{gf}$ & $\lambda_1$ & $\lambda_2$ & $\lambda_3$ \\
\midrule
\multirow{6}{*}{MuSHRoom} & \texttt{coffee\_room} & 600 & 0.08 & 0.1 & 0.08 & 0.4 \\
& \texttt{computer} & 600 & 0.5 & 0.1 & 0.08 & 0.4 \\
& \texttt{honka} & 600 & 0.1 & 0.1 & 0.08 & 0.4 \\
& \texttt{kokko} & 550 & 0.4 & 0.05 & 0.04 & 0.2 \\
& \texttt{sauna} & 550 & 0.3 & 0.025 & 0.02 & 0.1 \\
& \texttt{vr\_room} & 650 & 0.2 & 0.1 & 0.08 & 0.4 \\
\midrule
\multirow{2}{*}{ScanNet++} & \texttt{b20a261fdf} & 600 & 0.1 & 0.025 & 0.02 & 0.1 \\
& \texttt{8b5caf3398} & 600 & 0.1 & 0.025 & 0.02 & 0.1 \\
\bottomrule
\end{tabular}
\caption{Per-scene guided filter parameters and loss weights.}
\label{tab:hyperparams_appendix}
\end{table}

\subsection{ScanNet++ Setup}
We use the iPhone sequences of ScanNet++. Following the DN-Splatter data loader~\cite{turkulainen2025dn}, every tenth posed frame is held out for evaluation and every fifth remaining frame is used for training, which gives 59 training images for \texttt{b20a261fdf} and 126 for \texttt{8b5caf3398}. We train at half resolution ($960\times720$).

\subsection{Compute}
Our runs use a single NVIDIA RTX 3090 GPU. Training takes about 1.5 hours per MuSHRoom scene, and about 65 minutes per ScanNet++ scene at half resolution. The guided filter runs on the CPU (OpenCV).

\section{Definitions for Metrics}
\subsection{Depth Evaluation Metrics}
To rigorously assess depth estimation performance on the ScanNet++ and MuSHRoom datasets, we adopt standard evaluation protocols established by prior work~\cite{turkulainen2025dn}. Our quantitative assessment relies on the following metrics:

\begin{itemize}
    \item \textbf{Absolute Relative Error (Abs Rel):} Computes the mean ratio of the absolute depth difference to the corresponding ground-truth depth, reflecting the average magnitude of relative deviations across the scene.
    \item \textbf{Squared Relative Error (Sq Rel):} Similar to Abs Rel, but squares the relative depth differences before averaging, which more severely penalizes larger, localized depth errors.
    \item \textbf{Root Mean Squared Error (RMSE):} Calculates the square root of the mean squared differences between the predicted and actual depth maps, providing a direct measurement of the absolute error magnitude.
    \item \textbf{Logarithmic RMSE (RMSE log):} Computes the RMSE in the logarithmic domain. This metric is particularly advantageous for mitigating the disproportionate influence of distant background objects, which naturally exhibit larger depth values.
    \item \textbf{Threshold Accuracy ($\delta < t$):} Evaluates the percentage of predicted pixels where the maximum ratio between the predicted depth and ground-truth depth (or vice versa) falls below a predefined threshold factor $t$.
\end{itemize}

\subsection{Mesh Evaluation Metrics}
For 3D geometric fidelity, we compare our extracted 3D geometry against ground-truth meshes using widely accepted reconstruction metrics, with formal definitions provided in Table~\ref{tab:mesh_metrics_definitions}. To quantify the structural overlap between the predicted and actual geometry, we compute Precision, Recall, and the resulting F-score. Following standard practices, we enforce a strict distance threshold of 5~cm to determine whether a predicted vertex successfully matches the ground truth. Furthermore, to ensure a fair and relevant assessment, we strictly isolate our mesh quality evaluation to regions that fall within the visibility frustums of the training camera views.

\begin{table*}[h]
    \centering
    \renewcommand{\arraystretch}{1.6} 
    \resizebox{0.8\linewidth}{!}{
    \begin{tabular}{lc}
        \toprule
        \textbf{Metric} & \textbf{Mathematical Formulation} \\
        \midrule
        Accuracy & $\frac{1}{|Q|} \sum_{\mathbf{q} \in Q} \min_{\mathbf{g} \in G} \|\mathbf{q} - \mathbf{g}\|_1$ \\
        Completion & $\frac{1}{|G|} \sum_{\mathbf{g} \in G} \min_{\mathbf{q} \in Q} \|\mathbf{q} - \mathbf{g}\|_1$ \\
        Chamfer-$L_1$ & $\frac{1}{2} \left( \text{Accuracy} + \text{Completion} \right)$ \\
        \midrule
        Normal Accuracy & $\frac{1}{|Q|} \sum_{\mathbf{q} \in Q} \left( \mathbf{n}_{\mathbf{q}}^{\top} \mathbf{n}_{\mathbf{g}^*} \right)$ \quad s.t. $\mathbf{g}^* = \arg\min_{\mathbf{g} \in G} \|\mathbf{q} - \mathbf{g}\|_1$ \\
        Normal Completion & $\frac{1}{|G|} \sum_{\mathbf{g} \in G} \left( \mathbf{n}_{\mathbf{q}^*}^{\top} \mathbf{n}_{\mathbf{g}} \right)$ \quad s.t. $\mathbf{q}^* = \arg\min_{\mathbf{q} \in Q} \|\mathbf{q} - \mathbf{g}\|_1$ \\
        Normal Consistency & $\frac{1}{2} \left( \text{Normal Accuracy} + \text{Normal Completion} \right)$ \\
        \midrule
        Precision & $\frac{1}{|Q|} \sum_{\mathbf{q} \in Q} \mathbb{I} \left[ \min_{\mathbf{g} \in G} \|\mathbf{q} - \mathbf{g}\|_1 < \tau \right]$ \\
        Recall & $\frac{1}{|G|} \sum_{\mathbf{g} \in G} \mathbb{I} \left[ \min_{\mathbf{q} \in Q} \|\mathbf{q} - \mathbf{g}\|_1 < \tau \right]$ \\
        F-score & $\frac{2 \cdot \text{Precision} \cdot \text{Recall}}{\text{Precision} + \text{Recall}}$ \\
        \bottomrule
    \end{tabular}
    }
    \caption{\textbf{Formal Definitions of Mesh Evaluation Metrics.} We denote the point clouds sampled from the predicted mesh and the ground-truth mesh as $Q$ and $G$, respectively. For any given point $\mathbf{x}$, its corresponding surface normal vector is denoted as $\mathbf{n}_{\mathbf{x}}$. For classification metrics, $\mathbb{I}[\cdot]$ represents the indicator function, and the distance threshold is set to $\tau = 5$\,cm.}
    \label{tab:mesh_metrics_definitions}
\end{table*}

\section{Additional Qualitative Examples}
\subsection{Sparse-View Reconstruction} \label{sec:llff_appendix}
\begin{figure}[t]
  \centering
  \includegraphics[width=\textwidth]{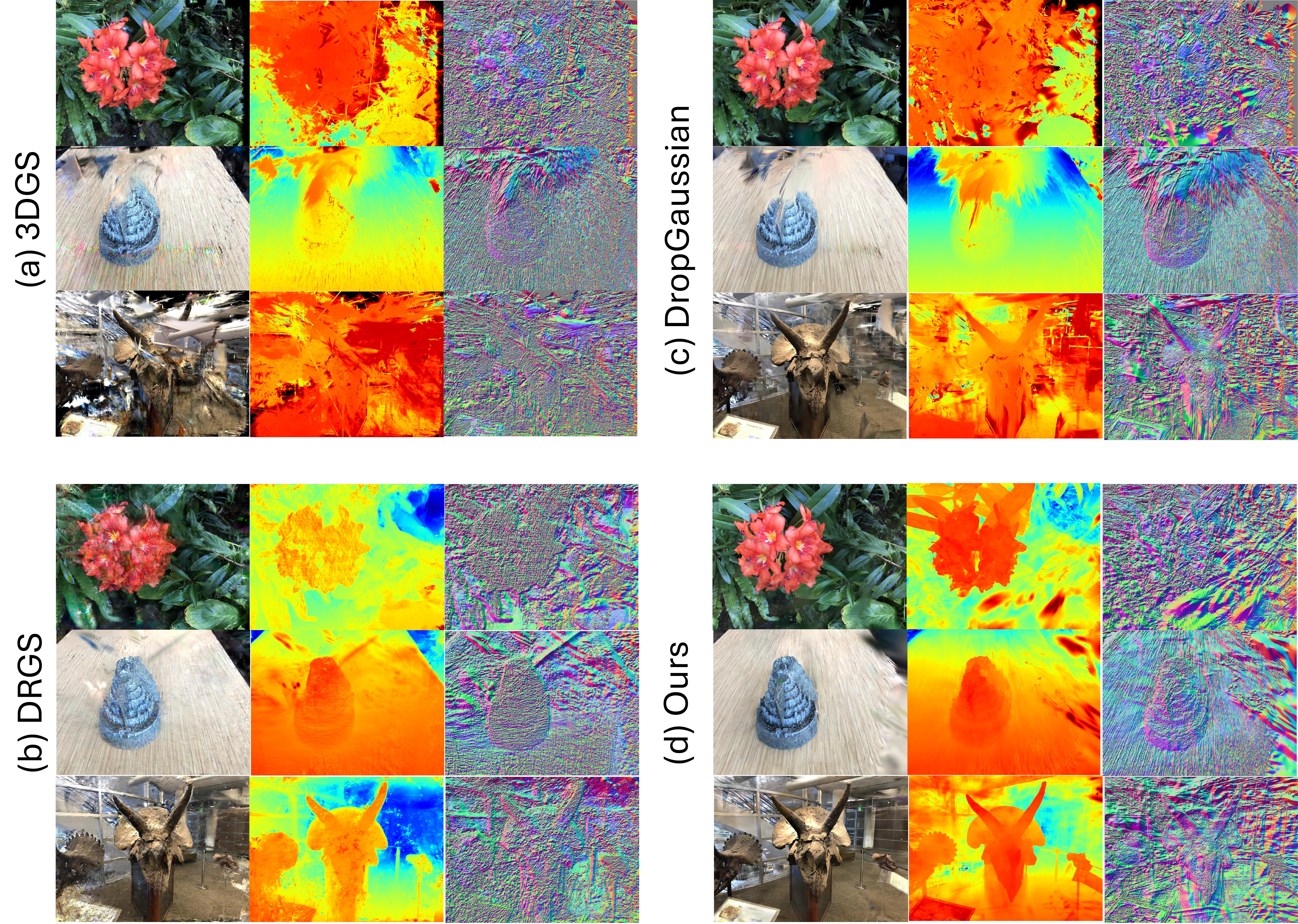}
  \caption{
  Additional qualitative comparison on NeRF-LLFF dataset. Visualized across the Flower, Fern, and Horns scenes under extreme view sparsity (rendered RGB, depth, and surface normal maps). (a) Unconstrained 3DGS severely overfits to the limited training views, resulting in massive floating artifacts and chaotic underlying geometry. (b) DRGS mitigates some artifacts via global depth alignment, but still suffers from noisy surface normals and blurred geometric details. (c) DropGaussian struggles to properly initialize in this sparse setting, leading to heavily corrupted backgrounds and incomplete structures. (d) In contrast, our proposed method leverages spatially rectified monocular priors to extract clean, highly accurate geometry with sharp object boundaries, smooth backgrounds, and highly consistent surface normals, all while preserving photorealistic appearance.
  }
  \label{fig:llff_appendix}
\end{figure}

We present additional qualitative comparisons for novel-view synthesis under highly constrained sparse-view settings on the NeRF-LLFF dataset in Fig.~\ref{fig:llff_appendix}. As observed in both the \texttt{Flower} and \texttt{Horns} scenes, the unconstrained 3DGS baseline severely overfits to the limited training views, resulting in massive clouds of floating artifacts and completely unstructured background geometry. While DropGaussian attempts to regularize the scene, its heavy reliance on an initial Structure-from-Motion (SfM) point cloud causes it to struggle under extreme sparsity, leading to incomplete object boundaries and persistent background noise. DRGS successfully reduces some floating artifacts through global depth alignment, but still exhibits geometric blurring and structural inaccuracies in complex regions. 

In contrast, our proposed method leverages spatially rectified monocular priors to continuously enforce strict geometric constraints throughout the optimization. This results in dramatically cleaner surface extraction, sharp object boundaries, and smooth, artifact-free backgrounds, perfectly corroborating the significant quantitative improvements reported in Table~\ref{tab:views_result} of the main text.

\subsection{Novel-View Synthesis and Depth Estimation} \label{sec:NVS_appendix}
\begin{figure}[t]
  \centering
  \includegraphics[width=\textwidth]{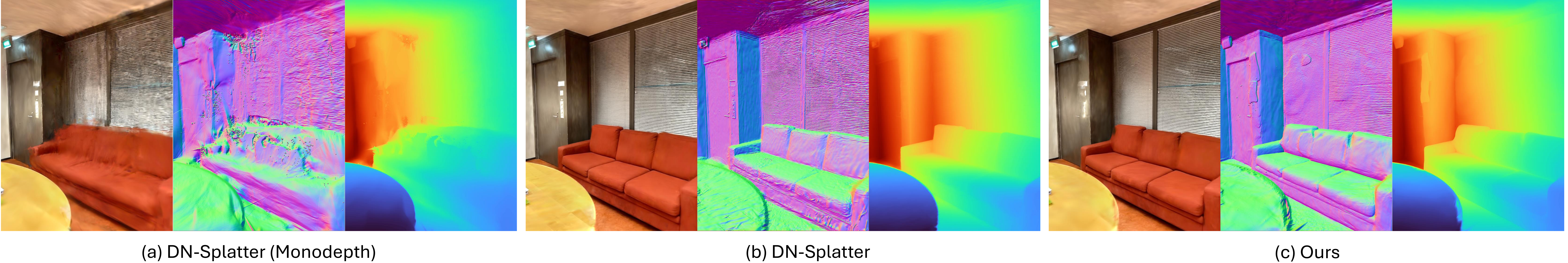}
  \caption{
  Additional qualitative comparison of rendered RGB, normals and depths on MuSHRoom dataset. While (a) the globally aligned DN-Splatter (Monodepth) baseline suffers from severe surface rippling and rendering artifacts, (c) our spatially rectified approach successfully recovers sharp, clean geometry that closely matches (b) the ground-truth sensor-supervised DN-Splatter.
  }
  \label{fig:mushroom_NVS_appendix}
\end{figure}

We present additional qualitative comparisons on the MuSHRoom dataset in Fig.~\ref{fig:mushroom_NVS_appendix}. We show that directly substituting physical sensor depth with uncorrected monocular priors leads to severe geometric degradation. The DN-Splatter (Monodepth) baseline, which relies on a single global scale-and-shift alignment, experiences catastrophic structural collapse. This is distinctly evident in the rendered normal maps: planar surfaces such as the back wall and the structure of the couch exhibit extreme geometric distortion. These underlying topological errors subsequently degrade the novel-view RGB rendering, resulting in blurred and warped textures. In contrast, our proposed method overcomes these limitations by applying spatially varying affine rectification to the exact same monocular prior. Notably, our sensor-free method reconstructs geometric and photometric fidelity that closely matches the original DN-Splatter, which possesses the massive advantage of physical sensor depth prior.


\subsection{Mesh Evaluation} \label{sec:mesh_appendix}
\begin{figure}[t]
  \centering
  \includegraphics[width=\textwidth]{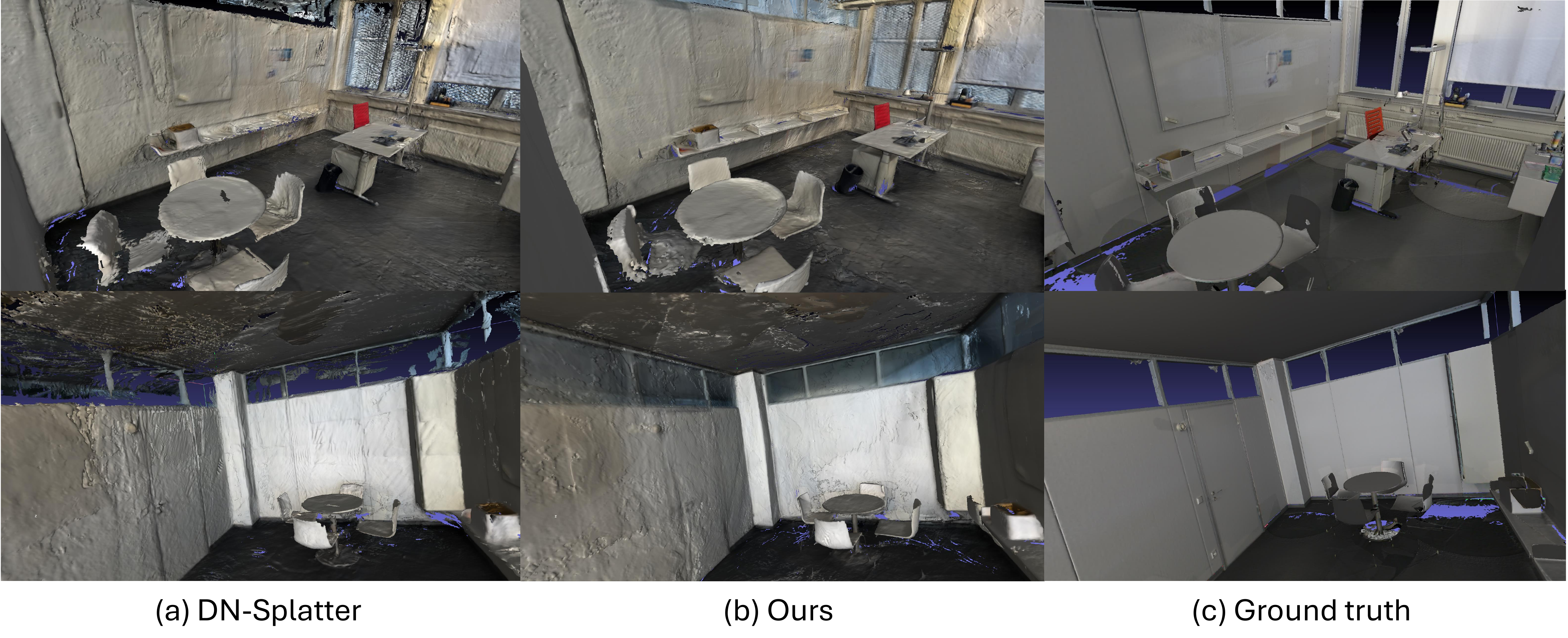}
  \caption{
  Additional qualitative mesh comparison on ScanNet++ dataset. (a) The DN-Splatter baseline exhibits high-frequency noise and artifacts across large planar surfaces like walls and ceilings. (b) Our proposed method successfully suppresses this noise to reconstruct much smoother, flatter surfaces that better approximate (c) the ground-truth geometry.
  }
  \label{fig:scannet_mesh_appendix}
\end{figure}

We provide additional qualitative comparisons for 3D mesh reconstruction on the ScanNet++ dataset in Fig.~\ref{fig:scannet_mesh_appendix}. We show that the mesh extracted by the DN-Splatter baseline exhibits noticeable high-frequency noise and irregular geometric artifacts, particularly across large planar surfaces such as walls and upper room structures. In contrast, our proposed method reconstructs significantly smoother and more consistent planar surfaces, visibly reducing these noisy geometric artifacts. 

\section{Additional Quantitative Results}

\subsection{Comprehensive Analysis of Geometric Regularizations} 
\begin{figure}[t]
  \centering
  \includegraphics[width=\textwidth]{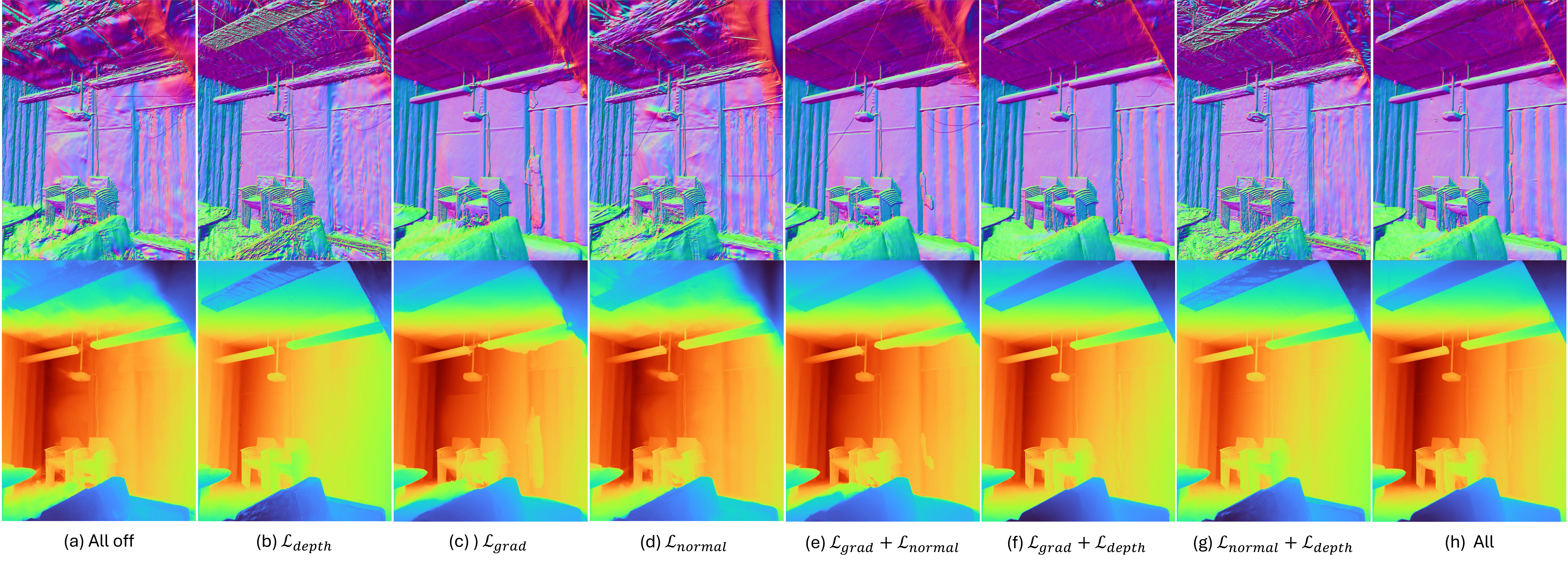}
  \caption{
    Qualitative comparisons of our proposed geometric regularization on MuSHRoom dataset. Visualized via rendered normals (top row) and depth maps (bottom row). \textbf{(a)} The unconstrained baseline (All off) exhibits severe geometric drift, yielding highly chaotic and noisy surface normals. \textbf{(b-d)} Applying individual loss terms in isolation yields incomplete geometry: while $\mathcal{L}_\text{depth}$ anchors global scale, it fails to resolve high-frequency structural noise, and while $\mathcal{L}_\text{grad}$ or $\mathcal{L}_\text{normal}$ improve local structure, they struggle without an absolute metric anchor. \textbf{(h)} Our full formulation combines these constraints. By anchoring the global metric scale while explicitly enforcing local spatial gradients and physical orientations, the complete approach reconstructs exceptionally clean planar surfaces (\eg, the walls and ceiling), sharp structural boundaries, and globally consistent depth.
  }
  \label{fig:ablation_loss_appendix}
\end{figure}

In Table~\ref{tab:full_loss_appendix}, we provide the complete quantitative ablation covering novel-view synthesis, depth estimation, and 3D mesh reconstruction metrics. The unconstrained Splatfacto baseline exhibits severe geometric drift, yielding a depth RMSE of 0.196 and a low Normal Consistency of 0.779. Introducing our metric depth re-alignment ($\mathcal{L}_\text{depth}$) directly addresses global scale ambiguity, evidenced by a sharp decrease in depth RMSE (to 0.147) and an increase in the inlier ratio $\delta < 1.25$ (from 0.937 to 0.984). However, global scale correction alone is insufficient for capturing high-frequency geometric details, resulting in a modest Normal Consistency of 0.793. Conversely, applying normal-aware supervision ($\mathcal{L}_\text{normal}$) or gradient-domain regularization ($\mathcal{L}_\text{grad}$) in isolation fails to anchor the global scene geometry; in fact, using solely $\mathcal{L}_\text{grad}$ degrades global depth metrics (Abs Rel of 0.085) because it only supervises relative spatial slopes without an absolute metric anchor. The true efficacy of our approach emerges when these complementary constraints are jointly applied. The $\mathcal{L}_\text{depth}$ term rigidly anchors the global metric scale, while $\mathcal{L}_\text{grad}$ and $\mathcal{L}_\text{normal}$ enforce precise local spatial gradients and physical surface orientations. This coupled formulation (All) achieves the lowest overall depth RMSE (0.139) and the highest Normal Consistency (0.891) and mesh F-score (0.855). This extensive evaluation confirms that our joint regularization successfully resolves the fundamental ambiguity between view-dependent appearance and underlying physical structure, exchanging a marginal drop in SSIM (0.869 vs.\ 0.858) for a highly robust, artifact-free 3D representation.  We present qualitative comparisons in Fig.~\ref{fig:ablation_loss_appendix}.
\begin{table*}[h]
    \centering
    \renewcommand{\arraystretch}{1.2}
    \resizebox{\textwidth}{!}{
    \begin{tabular}{l|ccc|cccc|ccccc}
        \toprule
        \multirow{2}{*}{\textbf{Method}} & \multicolumn{3}{c|}{\textbf{Photometry}} & \multicolumn{4}{c|}{\textbf{Depth Estimation}} & \multicolumn{5}{c}{\textbf{Mesh Reconstruction}} \\
        \cmidrule(lr){2-4} \cmidrule(lr){5-8} \cmidrule(lr){9-13}
        & PSNR $\uparrow$ & SSIM $\uparrow$ & LPIPS $\downarrow$ & Abs Rel $\downarrow$ & Sq Rel $\downarrow$ & RMSE $\downarrow$ & $\delta < 1.25 \uparrow$ & Acc. $\downarrow$ & Comp. $\downarrow$ & Chamfer-$L_1 \downarrow$ & Norm. Cons. $\uparrow$ & F-score $\uparrow$ \\
        \midrule
        Splatfacto & 25.293 & 0.869 & 0.086 & 0.060 & 0.021 & 0.196 & 0.937 & 0.043 & 0.039 & 0.041 & 0.779 & 0.728 \\
        \midrule
        + $\mathcal{L}_\text{grad}$ & 24.323 & 0.848 & 0.123 & 0.085 & 0.045 & 0.262 & 0.889 & 0.055 & 0.051 & 0.053 & 0.857 & 0.673 \\
        + $\mathcal{L}_\text{normal}$ & 25.257 & 0.867 & 0.085 & 0.064 & 0.022 & 0.205 & 0.926 & 0.043 & 0.039 & 0.041 & 0.797 & 0.728 \\
        + $\mathcal{L}_\text{depth}$ & 24.561 & 0.863 & 0.098 & 0.043 & 0.011 & 0.147 & 0.984 & 0.031 & 0.039 & 0.035 & 0.793 & 0.799 \\
        + $\mathcal{L}_\text{grad}$ + $\mathcal{L}_\text{normal}$ & 24.553 & 0.851 & 0.117 & 0.075 & 0.034 & 0.238 & 0.902 & 0.051 & 0.047 & 0.049 & 0.868 & 0.707 \\
        + $\mathcal{L}_\text{grad}$ + $\mathcal{L}_\text{depth}$ & 25.013 & 0.860 & 0.101 & 0.041 & 0.010 & 0.141 & 0.983 & \textbf{0.026} & \textbf{0.028} & \textbf{0.027} & 0.885 & 0.850 \\
        + $\mathcal{L}_\text{normal}$ + $\mathcal{L}_\text{depth}$ & \textbf{25.419} & \textbf{0.869} & \textbf{0.084} & \textbf{0.040} & \textbf{0.010} & 0.141 & \textbf{0.987} & 0.029 & 0.032 & 0.031 & 0.819 & 0.814 \\
        \midrule
        \textbf{All (Ours)} & 24.947 & 0.858 & 0.103 & \textbf{0.040} & \textbf{0.010} & \textbf{0.139} & 0.985 & \textbf{0.026} & \textbf{0.028} & \textbf{0.027} & \textbf{0.891} & \textbf{0.855} \\
        \bottomrule
    \end{tabular}
    }
    \caption{Comprehensive ablations on proposed geometric regularization components. Evaluated on the \texttt{coffee\_room} scene from the MuSHRoom dataset. Full quantitative evaluation of the individual and combined contributions of our proposed geometric loss terms. We report photometric (novel-view synthesis), depth estimation, and 3D mesh reconstruction metrics against the unconstrained Splatfacto baseline. Our full configuration (\textit{All}) provides the best overall balance, enforcing strict structural integrity (highest mesh F-score and Normal Consistency) while maintaining competitive rendering quality. Best results in each column are indicated in \textbf{bold}.}
    \label{tab:full_loss_appendix}
\end{table*}

\subsection{Sensitivity to Monocular Depth Priors} \label{sec:mono_prior_ablation_appendix}

In Table~\ref{tab:full_depth_prior_appendix}, we provide the complete scene-wise quantitative evaluation of our method across three state-of-the-art monocular depth models: Depth-Anything-V2~\cite{yang2024depth}, MoGe~\cite{wang2025moge}, and Metric3D-V2~\cite{hu2024metric3dv2}. While the average metrics demonstrate the general robustness of our approach, the scene-by-scene breakdown reveals critical insights into prior network stability. Both Depth-Anything-V2 and MoGe demonstrate exceptional consistency across diverse indoor topologies. For instance, in the highly textured \texttt{coffee\_room} scene, both models facilitate outstanding mesh F-scores (0.855 and 0.879, respectively) and near-perfect inlier ratios ($\delta < 1.25$ over 0.98). 

Conversely, the performance of Metric3D-V2 exhibits high variance, largely due to an architectural constraint: Metric3D-V2 utilizes a fixed wide-aspect-ratio inference canvas. Because the MuSHRoom dataset consists of portrait-oriented captures, feeding these images into the network requires massive aspect-ratio padding, which systematically induces severe depth and disparity artifacts near the image borders. Consequently, in challenging environments like the \texttt{computer} and \texttt{vr\_room} scenes—where these border artifacts intersect with thin structures and complex occlusion boundaries—Metric3D-V2 suffers catastrophic structural degradation, yielding F-scores of just 0.428 and 0.458, respectively. This scene-specific collapse heavily penalizes its average performance. These comprehensive results validate that while our spatially varying affine rectification successfully bridges the gap between relative and metric depth, the ultimate geometric fidelity of the 3DGS representation remains bounded by the structural reliability and boundary-awareness of the chosen foundation prior.

\begin{table*}[h]
    \centering
    \renewcommand{\arraystretch}{1.1}
    \resizebox{\textwidth}{!}{
    \begin{tabular}{ll|ccc|cccc|ccccc}
        \toprule
        \multirow{2}{*}{\textbf{Depth Prior}} & \multirow{2}{*}{\textbf{Scene}} & \multicolumn{3}{c|}{\textbf{Photometry}} & \multicolumn{4}{c|}{\textbf{Depth Estimation}} & \multicolumn{5}{c}{\textbf{Mesh Reconstruction}} \\
        \cmidrule(lr){3-5} \cmidrule(lr){6-9} \cmidrule(lr){10-14}
        & & PSNR $\uparrow$ & SSIM $\uparrow$ & LPIPS $\downarrow$ & Abs Rel $\downarrow$ & Sq Rel $\downarrow$ & RMSE $\downarrow$ & $\delta < 1.25 \uparrow$ & Acc. $\downarrow$ & Comp. $\downarrow$ & Chamfer-$L_1 \downarrow$ & Norm. Cons. $\uparrow$ & F-score $\uparrow$ \\
        \midrule
        \multirow{7}{*}{\textbf{Depth-Anything-V2}} 
        & coffee\_room & 24.947 & 0.858 & 0.103 & 0.040 & 0.010 & 0.139 & 0.985 & 0.026 & 0.028 & 0.027 & 0.891 & 0.855 \\
        & computer & 24.968 & 0.851 & 0.129 & 0.070 & 0.026 & 0.254 & 0.955 & 0.065 & 0.058 & 0.061 & 0.884 & 0.603 \\
        & honka & 23.142 & 0.821 & 0.140 & 0.050 & 0.017 & 0.175 & 0.973 & 0.047 & 0.034 & 0.041 & 0.885 & 0.799 \\
        & kokko & 22.444 & 0.831 & 0.163 & 0.064 & 0.025 & 0.220 & 0.933 & 0.051 & 0.042 & 0.047 & 0.798 & 0.705 \\
        & sauna & 24.443 & 0.830 & 0.187 & 0.060 & 0.015 & 0.167 & 0.978 & 0.051 & 0.044 & 0.047 & 0.874 & 0.687 \\
        & vr\_room & 26.144 & 0.855 & 0.163 & 0.045 & 0.013 & 0.153 & 0.981 & 0.081 & 0.057 & 0.069 & 0.866 & 0.608 \\
        \cmidrule(lr){2-14}
        & \textbf{Average} & 24.348 & 0.841 & 0.148 & 0.055 & 0.018 & 0.185 & 0.968 & 0.054 & 0.044 & 0.049 & 0.866 & 0.710 \\
        \midrule
        \multirow{7}{*}{\textbf{MoGe}} 
        & coffee\_room & 25.088 & 0.863 & 0.097 & 0.039 & 0.010 & 0.139 & 0.984 & 0.026 & 0.026 & 0.026 & 0.894 & 0.879 \\
        & computer & 24.569 & 0.848 & 0.137 & 0.091 & 0.038 & 0.293 & 0.915 & 0.074 & 0.083 & 0.079 & 0.860 & 0.626 \\
        & honka & 23.171 & 0.820 & 0.144 & 0.051 & 0.016 & 0.166 & 0.973 & 0.035 & 0.030 & 0.033 & 0.886 & 0.822 \\
        & kokko & 22.197 & 0.825 & 0.163 & 0.072 & 0.031 & 0.246 & 0.913 & 0.054 & 0.048 & 0.051 & 0.793 & 0.675 \\
        & sauna & 24.445 & 0.828 & 0.179 & 0.066 & 0.017 & 0.178 & 0.967 & 0.047 & 0.045 & 0.046 & 0.874 & 0.687 \\
        & vr\_room & 26.301 & 0.861 & 0.150 & 0.046 & 0.014 & 0.164 & 0.977 & 0.062 & 0.050 & 0.056 & 0.883 & 0.649 \\
        \cmidrule(lr){2-14}
        & \textbf{Average} & 24.295 & 0.841 & 0.145 & 0.061 & 0.021 & 0.198 & 0.955 & 0.050 & 0.047 & 0.048 & 0.865 & 0.723 \\
        \midrule
        \multirow{7}{*}{\textbf{Metric3D-V2}} 
        & coffee\_room & 24.429 & 0.850 & 0.117 & 0.053 & 0.015 & 0.177 & 0.958 & 0.057 & 0.056 & 0.056 & 0.845 & 0.673 \\
        & computer & 21.819 & 0.804 & 0.225 & 0.237 & 0.215 & 0.621 & 0.626 & 0.190 & 0.261 & 0.226 & 0.763 & 0.428 \\
        & honka & 22.696 & 0.815 & 0.157 & 0.068 & 0.022 & 0.204 & 0.937 & 0.058 & 0.060 & 0.059 & 0.849 & 0.673 \\
        & kokko & 21.797 & 0.812 & 0.187 & 0.095 & 0.053 & 0.320 & 0.877 & 0.106 & 0.074 & 0.090 & 0.749 & 0.508 \\
        & sauna & 23.102 & 0.796 & 0.241 & 0.150 & 0.136 & 0.380 & 0.811 & 0.087 & 0.103 & 0.095 & 0.831 & 0.556 \\
        & vr\_room & 25.518 & 0.847 & 0.180 & 0.075 & 0.029 & 0.242 & 0.907 & 0.131 & 0.101 & 0.116 & 0.799 & 0.458 \\
        \cmidrule(lr){2-14}
        & \textbf{Average} & 23.227 & 0.821 & 0.184 & 0.113 & 0.078 & 0.324 & 0.853 & 0.105 & 0.109 & 0.107 & 0.806 & 0.549 \\
        \bottomrule
    \end{tabular}
    }
    \caption{Ablation on monocular depth models. Evaluated across all six scenes of the MuSHRoom dataset test split. We report full photometric, depth estimation, and 3D mesh reconstruction metrics for our proposed method when supervised by three different foundation models (Depth-Anything-V2, MoGe, and Metric3D-V2). While our affine rectification effectively extracts metric geometry regardless of the prior, high-quality estimators like MoGe and Depth-Anything-V2 maintain significantly higher consistency across challenging scenes.}
    \label{tab:full_depth_prior_appendix}
\end{table*}

\subsection{Sensitivity to Filters}\label{sec:ablation_filter_appendix}
Raw outputs from monocular depth networks, while structurally strong, inevitably contain local noise, quantization artifacts, and high-frequency inconsistencies. If injected directly into the optimization pipeline without regularization, these noisy coefficients force the 3D Gaussians to overfit to erroneous local depth shifts, severely degrading both the extracted geometry and the novel view synthesis. To address this, we apply a filtering mechanism to smooth the extracted alignment coefficients. In this section, we ablate our choice of the Guided Filter against a standard Gaussian filter and a baseline with no filtering.

In Table~\ref{tab:ablation_filter_appendix}, we perform this ablation on the \texttt{coffee\_room} scene from the MuSHRoom dataset using the Depth-Anything-V2 prior. We keep all loss terms active (RGB + Disparity Normal Gradient + $L_1$ Depth) and isolate the effect of the filter type on the disparity alignment coefficients. The results are evaluated across Novel View Synthesis (\texttt{right} split), Depth estimation (\texttt{right} split), and Mesh extraction.

We show that disabling the filter entirely (``None'') results in a catastrophic drop in geometric accuracy; the F-score collapses to 0.680, and the depth RMSE spikes to 0.269, proving that unregularized monocular noise actively harms the spatial reconstruction. While applying a standard Gaussian blur mitigates some of this noise and improves the F-score to 0.774, it does so indiscriminately. Because a Gaussian filter smooths across physical depth boundaries, it inadvertently blurs distinct structural edges, limiting the sharpness of the final geometry. In contrast, our chosen Guided Blur leverages the prior itself as an edge-preserving guide. This ensures that noise is smoothed across flat surfaces while sharp structural discontinuities are strictly preserved. As a result, our Guided Blur strategy firmly dominates across all categories, yielding the highest photometric fidelity, the lowest depth error, and a significantly superior geometric F-score of 0.855.
\begin{table*}[h]
\centering
\begin{adjustbox}{max width=1.0\textwidth}
\setlength{\tabcolsep}{4pt}
\renewcommand{\arraystretch}{1.05}
\begin{tabular}{l | ccc | ccc | ccccc}
\toprule
\multirow{2}{*}{\textbf{Filter Type}} & \multicolumn{3}{c|}{\textbf{Novel View Synthesis}} & \multicolumn{3}{c|}{\textbf{Depth Evaluation}} & \multicolumn{5}{c}{\textbf{Mesh Evaluation}} \\
\cmidrule(lr){2-4} \cmidrule(lr){5-7} \cmidrule(lr){8-12}
& \textbf{PSNR}~$\uparrow$ & \textbf{SSIM}~$\uparrow$ & \textbf{LPIPS}~$\downarrow$ & \textbf{Abs Rel}~$\downarrow$ & \textbf{RMSE}~$\downarrow$ & \textbf{$\delta_1$}~$\uparrow$ & \textbf{Accuracy}~$\downarrow$ & \textbf{Comp.}~$\downarrow$ & \textbf{Chamfer-$L_1$}~$\downarrow$ & \textbf{Norm. Cons.}~$\uparrow$ & \textbf{F-score}~$\uparrow$ \\
\midrule
None & 20.66 & 0.767 & 0.231 & 0.080 & 0.269 & 0.858 & 0.0530 & 0.0530 & 0.0551 & 0.8592 & 0.6804 \\
Gaussian Blur & 21.29 & 0.770 & 0.206 & 0.038 & 0.125 & 0.986 & 0.0338 & 0.0347 & 0.0343 & 0.8786 & 0.7737 \\
\rowcolor{gray!10} Guided Blur (Ours) & \textbf{21.53} & \textbf{0.787} & \textbf{0.195} & \textbf{0.031} & \textbf{0.112} & \textbf{0.991} & \textbf{0.0258} & \textbf{0.0278} & \textbf{0.0268} & \textbf{0.8908} & \textbf{0.8551} \\
\bottomrule
\end{tabular}
\end{adjustbox}
\vspace{0.1cm}
\caption{Ablation of prior filtering strategies. Evaluated on the \texttt{coffee\_room} scene using the Depth-Anything-V2 prior. Novel-view synthesis and depth metrics are reported on the \texttt{right} (disjoint-trajectory) split, whereas Tables~\ref{tab:loss_ablation} and~\ref{tab:full_depth_prior_appendix} report the \texttt{left} split. Removing the filter results in severe geometric degradation due to raw monocular noise. While a standard Gaussian filter mitigates some noise, it indiscriminately blurs across physical boundaries. Our Guided Blur strategy uniquely preserves sharp structural edges, yielding the best performance across all photometric, depth, and geometric metrics.}
\label{tab:ablation_filter_appendix}
\end{table*}

\subsection{Sensitivity to Gradient Matching Formulation}\label{sec:ablation_gradient_formulation_appendix}
We evaluate the sensitivity of our approach to the specific mathematical formulation of the disparity  gradient loss defined in Eq.~\ref{eq:dispariy_gradient}. Using the \texttt{coffee\_room} scene and the Depth-Anything-V2 prior, we ablate our combined magnitude and directional alignment formulation (MSE + Cosine) against using strictly Mean Squared Error (MSE) or strictly Cosine Similarity. The objective of this loss term is to ensure that the spatial gradients of the rendered depth map closely follow the structural gradients of the monocular prior. 

The quantitative results in Table~\ref{tab:ablation_gradient_appendix} demonstrate that relying exclusively on an MSE formulation is insufficient for complex geometries. MSE alone struggles to enforce sharp orientational changes, resulting in a significant drop in Normal Consistency (0.791) and overall F-score (0.781). While Cosine Similarity alone provides a highly effective directional cue—matching our combined approach in absolute depth accuracy—it lacks a constraint on the actual magnitude of the gradient change. By combining both MSE (magnitude) and Cosine (direction), we ensure the most robust structural regularization. This combined approach prevents local geometric collapse and yields the best overall surface extraction (F-score of 0.855) alongside optimal photometric rendering quality (21.53 dB PSNR).

\begin{table*}[h]
\centering
\begin{adjustbox}{max width=1.0\textwidth}
\setlength{\tabcolsep}{4pt}
\renewcommand{\arraystretch}{1.05}
\begin{tabular}{l | ccc | ccc | ccccc}
\toprule
\multirow{2}{*}{\textbf{Gradient Loss}} & \multicolumn{3}{c|}{\textbf{Novel View Synthesis}} & \multicolumn{3}{c|}{\textbf{Depth Evaluation}} & \multicolumn{5}{c}{\textbf{Mesh Evaluation}} \\
\cmidrule(lr){2-4} \cmidrule(lr){5-7} \cmidrule(lr){8-12}
& \textbf{PSNR}~$\uparrow$ & \textbf{SSIM}~$\uparrow$ & \textbf{LPIPS}~$\downarrow$ & \textbf{Abs Rel}~$\downarrow$ & \textbf{RMSE}~$\downarrow$ & \textbf{$\delta_1$}~$\uparrow$ & \textbf{Accuracy}~$\downarrow$ & \textbf{Comp.}~$\downarrow$ & \textbf{Chamfer-$L_1$}~$\downarrow$ & \textbf{Norm. Cons.}~$\uparrow$ & \textbf{F-score}~$\uparrow$ \\
\midrule
MSE & 20.39 & 0.776 & 0.226 & 0.055 & 0.157 & 0.959 & 0.0328 & 0.0400 & 0.0364 & 0.7914 & 0.7815 \\
Cosine & 21.44 & 0.786 & \textbf{0.195} & \textbf{0.030} & \textbf{0.110} & \textbf{0.991} & 0.0259 & \textbf{0.0278} & 0.0269 & \textbf{0.8911} & 0.8535 \\
\rowcolor{gray!10} MSE + Cosine (Ours) & \textbf{21.53} & \textbf{0.787} & \textbf{0.195} & 0.031 & 0.112 & \textbf{0.991} & \textbf{0.0258} & \textbf{0.0278} & \textbf{0.0268} & 0.8908 & \textbf{0.8551} \\
\bottomrule
\end{tabular}
\end{adjustbox}
\vspace{0.1cm}
\caption{Ablation on gradient matching formulation. Evaluated on the \texttt{coffee\_room} scene. Novel-view synthesis and depth metrics are reported on the \texttt{right} (disjoint-trajectory) split. Relying solely on MSE fails to capture sharp orientational changes, leading to degraded Normal Consistency. While Cosine similarity provides excellent directional cues, combining both magnitude (MSE) and directional (Cosine) alignment yields the highest overall surface extraction F-score and best novel view synthesis.}
\label{tab:ablation_gradient_appendix}
\end{table*}

\subsection{Effect of the Meshing Algorithm} \label{sec:meshing_appendix}
Mesh metrics depend strongly on the algorithm used to extract the mesh. Table~\ref{tab:meshing_appendix} compares Poisson reconstruction and TSDF fusion applied to the same trained models on MuSHRoom. TSDF fusion gives better meshes for all three methods on every metric; for DN-Splatter with sensor depth, it halves the Chamfer distance and raises the F-score by 0.11. Comparisons between methods in Tables~\ref{tab:main-table-mesh-mushroom} and~\ref{tab:main-table-mesh-scannetpp} should therefore be read together with their Algorithm column. The numbers in Table~\ref{tab:meshing_appendix} come from our reruns with the released code, so they differ slightly from those in Table~\ref{tab:main-table-mesh-mushroom}.

\begin{table}[h]
\centering
\begin{adjustbox}{max width=\textwidth}
\begin{tabular}{llccccc}
\toprule
Method & Algorithm & Accuracy~$\downarrow$ & Completion~$\downarrow$ & Chamfer-$L_1$~$\downarrow$ & Normal Cons.~$\uparrow$ & F-score~$\uparrow$ \\
\midrule
\multirow{2}{*}{DN-Splatter} & Poisson & .0405 & .0421 & .0413 & .8322 & .8299 \\
& TSDF & .0225 & .0202 & .0213 & .8847 & .9381 \\
\midrule
\multirow{2}{*}{DN-Splatter (Monodepth)} & Poisson & .2204 & .4045 & .3125 & .6778 & .2280 \\
& TSDF & .2162 & .3397 & .2780 & .7095 & .2769 \\
\midrule
\multirow{2}{*}{Ours$^\dagger$} & Poisson & .0723 & .0656 & .0690 & .8197 & .5883 \\
& TSDF & .0607 & .0445 & .0526 & .8795 & .7117 \\
\bottomrule
\end{tabular}
\end{adjustbox}
\caption{Poisson reconstruction versus TSDF fusion on the same trained models (MuSHRoom). DN-Splatter and DN-Splatter (Monodepth) are averaged over all six scenes. $^\dagger$Averaged over \texttt{coffee\_room}, \texttt{computer}, \texttt{honka} and \texttt{vr\_room}.}
\label{tab:meshing_appendix}
\end{table}

\section{Baselines}
\subsection{Comprehensive Analysis of the DN-Splatter (Monodepth) Baseline}
\label{sec:appendix_dnsplatter_analysis}

To fully demonstrate the limitations of global alignment strategies when utilizing monocular depth priors, we provide a comprehensive scene-wise breakdown of the DN-Splatter (Monodepth) baseline on the MuSHRoom dataset (Table~\ref{tab:full_dnsplatter_appendix}). While DN-Splatter yields exceptional results when provided with ground-truth sensor depth, its geometric integrity collapses when forced to rely on uncorrected monocular priors using a global scale-and-shift alignment. 

The scene-wise metrics reveal high volatility and catastrophic failures in structurally complex environments. While simpler or more uniformly constrained scenes like \texttt{coffee\_room} and \texttt{honka} manage to retain a loose semblance of geometry (F-scores of 0.516 and 0.409, respectively), environments with high depth variance and complex occlusion boundaries suffer from complete topological collapse. Specifically, in the \texttt{computer} and \texttt{kokko} scenes, the baseline yields extreme depth errors (RMSE $> 1.95$) and effectively zero structural reconstruction, plummeting to F-scores of 0.065 and 0.027. This erratic behavior perfectly illustrates the piecewise-linear fragmentation of foundation depth models discussed in our main text. Because monocular distortions vary spatially, forcing a single global linear alignment causes severe warping—optimizing to fit one local region inherently breaks the geometry of the rest of the scene. These results empirically validate our core motivation: to successfully harness monocular priors for 3DGS, spatially varying local rectification is strictly required.

\begin{table*}[h]
    \centering
    \renewcommand{\arraystretch}{1.1}
    \resizebox{\textwidth}{!}{
    \begin{tabular}{l|ccc|cccc|ccccc}
        \toprule
        \multirow{2}{*}{\textbf{Scene}} & \multicolumn{3}{c|}{\textbf{Photometry (Left / Right)}} & \multicolumn{4}{c|}{\textbf{Depth Estimation (Left / Right)}} & \multicolumn{5}{c}{\textbf{Mesh Reconstruction}} \\
        \cmidrule(lr){2-4} \cmidrule(lr){5-8} \cmidrule(lr){9-13}
        & PSNR $\uparrow$ & SSIM $\uparrow$ & LPIPS $\downarrow$ & Abs Rel $\downarrow$ & Sq Rel $\downarrow$ & RMSE $\downarrow$ & $\delta < 1.25 \uparrow$ & Acc. $\downarrow$ & Comp. $\downarrow$ & Chamfer-$L_1 \downarrow$ & Norm. Cons. $\uparrow$ & F-score $\uparrow$ \\
        \midrule
        coffee\_room & 22.71 / 19.74 & 0.788 / 0.727 & 0.187 / 0.288 & 0.088 / 0.086 & 0.033 / 0.033 & 0.245 / 0.232 & 0.862 / 0.903 & 0.076 & 0.084 & 0.080 & 0.792 & 0.516 \\
        computer     & 13.60 / 13.27 & 0.637 / 0.602 & 0.495 / 0.506 & 0.744 / 0.635 & 1.791 / 1.875 & 2.273 / 2.541 & 0.100 / 0.242 & 0.276 & 0.666 & 0.471 & 0.641 & 0.065 \\
        honka        & 21.38 / 19.40 & 0.755 / 0.705 & 0.218 / 0.272 & 0.101 / 0.091 & 0.047 / 0.028 & 0.275 / 0.244 & 0.877 / 0.926 & 0.124 & 0.109 & 0.117 & 0.811 & 0.409 \\
        kokko        & 12.81 / 11.38 & 0.617 / 0.592 & 0.537 / 0.585 & 0.823 / 0.921 & 1.649 / 2.256 & 1.958 / 2.532 & 0.046 / 0.009 & 0.445 & 0.689 & 0.567 & 0.524 & 0.027 \\
        sauna        & 20.95 / 18.52 & 0.745 / 0.718 & 0.329 / 0.373 & 0.284 / 0.356 & 0.319 / 0.443 & 0.596 / 0.708 & 0.619 / 0.545 & 0.136 & 0.240 & 0.188 & 0.724 & 0.316 \\
        vr\_room     & 24.12 / 21.16 & 0.811 / 0.763 & 0.236 / 0.288 & 0.085 / 0.113 & 0.034 / 0.076 & 0.264 / 0.358 & 0.906 / 0.866 & 0.151 & 0.129 & 0.140 & 0.764 & 0.326 \\
        \midrule
        \textbf{Average} & 19.26 / 17.24 & 0.725 / 0.684 & 0.334 / 0.385 & 0.354 / 0.367 & 0.645 / 0.785 & 0.935 / 1.103 & 0.568 / 0.582 & 0.201 & 0.320 & 0.261 & 0.709 & 0.277 \\
        \bottomrule
    \end{tabular}
    }
    \caption{Comprehensive scene-wise evaluation of the DN-Splatter (Monodepth) baseline. Evaluated across the MuSHRoom dataset test split. Because this baseline relies on a single global scale-and-shift transform to align monocular depth, it is highly susceptible to piecewise-linear distortions. This results in severe structural collapse on complex scenes (e.g., \texttt{computer} and \texttt{kokko}), validating the necessity of our proposed spatially varying affine rectification.}
    \label{tab:full_dnsplatter_appendix}
\end{table*}

\section{Limitations}
\label{sec:limitations}

While our proposed spatially varying affine rectification significantly improves the geometric fidelity of 3DGS, our approach has several limitations:

\textbf{Dependency on Foundation Model Reliability.} Because our method relies on extracting geometric cues from monocular depth estimators, our reconstruction quality is inherently bounded by the reliability of the chosen foundation model. As discussed in Appendix Sec.~\ref{sec:mono_prior_ablation_appendix}, if the monocular prior completely fails to capture the underlying structure—such as in the presence of highly transparent surfaces, massive specular reflections (mirrors), or severe domain gaps not covered by the prior's training data—our affine rectification cannot recover the lost geometry. 

\textbf{Absence of Absolute Physical Scale.} Our method bridges the gap toward sensor-supervised geometric fidelity by enforcing consistent Euclidean geometry across the scene. However, because it operates entirely without physical sensor data, the final reconstructed mesh still lacks an absolute real-world physical scale. The global metric anchor ($\mathcal{L}_\text{depth}$) aligns the geometry to the arbitrary coordinate space established by the initial SfM camera poses, rather than true physical dimensions. Moreover, because $\mathcal{L}_\text{depth}$ is re-aligned to the current rendering at every step, it cannot correct a global scale error that is already present in the photometric reconstruction. On the ScanNet++ scene \texttt{b20a261fdf}, both plain Splatfacto and our method render depth about 5\% shorter than the sensor depth. At the 2--4\,m viewing distances of this scene, this amounts to 10--20\,cm, beyond the 5\,cm F-score threshold, which limits our F-score on this scene to 0.18, compared with 0.46 on \texttt{8b5caf3398}.

\textbf{Sensitivity to Regularization Strength.} The default loss weights are too strong for some scenes. On \texttt{kokko} and \texttt{sauna}, they produce floaters, including directly in front of the camera, and we reduce all three weights by factors of 0.5 and 0.25, respectively (Appendix Sec.~\ref{sec:hyperparams_appendix}).


\clearpage
\section*{NeurIPS Paper Checklist}

\begin{enumerate}

\item {\bf Claims}
    \item[] Question: Do the main claims made in the abstract and introduction accurately reflect the paper's contributions and scope?
    \item[] Answer: \answerYes{} 
    \item[] Justification: The claims are empirically verified in Sec.~\ref{sec:experiments}.

\item {\bf Limitations}
    \item[] Question: Does the paper discuss the limitations of the work performed by the authors?
    \item[] Answer: \answerYes{} 
    \item[] Justification: The limitations are discussed in Appendix Sec.~\ref{sec:limitations}.

\item {\bf Theory assumptions and proofs}
    \item[] Question: For each theoretical result, does the paper provide the full set of assumptions and a complete (and correct) proof?
    \item[] Answer: \answerYes{} 
    \item[] Justification: We present full derivations in Appendix Sec.~\ref{supp:derivation}.

    \item {\bf Experimental result reproducibility}
    \item[] Question: Does the paper fully disclose all the information needed to reproduce the main experimental results of the paper to the extent that it affects the main claims and/or conclusions of the paper (regardless of whether the code and data are provided or not)?
    \item[] Answer: \answerYes{} 
    \item[] Justification: We disclose all implementation details in Sec.~\ref{sec:experiments}, and the per-scene settings in Appendix Sec.~\ref{sec:hyperparams_appendix}.

\item {\bf Open access to data and code}
    \item[] Question: Does the paper provide open access to the data and code, with sufficient instructions to faithfully reproduce the main experimental results, as described in supplemental material?
    \item[] Answer: \answerYes{} 
    \item[] Justification: We will provide code and instructions for the final paper.

\item {\bf Experimental setting/details}
    \item[] Question: Does the paper specify all the training and test details (e.g., data splits, hyperparameters, how they were chosen, type of optimizer) necessary to understand the results?
    \item[] Answer: \answerYes{}
    \item[] Justification: We provide complete experimental details in Sec.~\ref{sec:experiments} and Appendix Sec.~\ref{sec:hyperparams_appendix}.

\item {\bf Experiment statistical significance}
    \item[] Question: Does the paper report error bars suitably and correctly defined or other appropriate information about the statistical significance of the experiments?
    \item[] Answer: \answerNo{} 
    \item[] Justification: While we do not report explicit error bars or standard deviations to maintain table readability, we ensure the statistical reliability of our most sensitive experiments (the sparse-view evaluations) by running and averaging all reported metrics over 10 random seed selections (as detailed in Sec.~\ref{sec:experiments}).

\item {\bf Experiments compute resources}
    \item[] Question: For each experiment, does the paper provide sufficient information on the computer resources (type of compute workers, memory, time of execution) needed to reproduce the experiments?
    \item[] Answer: \answerYes{} 
    \item[] Justification: We report the GPU and training times in Appendix Sec.~\ref{sec:hyperparams_appendix}.
    
\item {\bf Code of ethics}
    \item[] Question: Does the research conducted in the paper conform, in every respect, with the NeurIPS Code of Ethics \url{https://neurips.cc/public/EthicsGuidelines}?
    \item[] Answer: \answerYes{} 
    \item[] Justification: We have read and understood the code of ethics and have made every effort to adhere to it.

\item {\bf Broader impacts}
    \item[] Question: Does the paper discuss both potential positive societal impacts and negative societal impacts of the work performed?
    \item[] Answer: \answerNA{} 
    \item[] Justification: There is no societal impact of the work performed.
    
\item {\bf Safeguards}
    \item[] Question: Does the paper describe safeguards that have been put in place for responsible release of data or models that have a high risk for misuse (e.g., pre-trained language models, image generators, or scraped datasets)?
    \item[] Answer: \answerNA{} 
    \item[] Justification: We do not release data or models so there is no such risks.

\item {\bf Licenses for existing assets}
    \item[] Question: Are the creators or original owners of assets (e.g., code, data, models), used in the paper, properly credited and are the license and terms of use explicitly mentioned and properly respected?
    \item[] Answer: \answerYes{} 
    \item[] Justification: We use publicly available dataset and pretrained models where all of them are properly cited in the paper. 

\item {\bf New assets}
    \item[] Question: Are new assets introduced in the paper well documented and is the documentation provided alongside the assets?
    \item[] Answer: \answerYes{} 
    \item[] Justification: We will release our codebase with detailed README files. 

\item {\bf Crowdsourcing and research with human subjects}
    \item[] Question: For crowdsourcing experiments and research with human subjects, does the paper include the full text of instructions given to participants and screenshots, if applicable, as well as details about compensation (if any)? 
    \item[] Answer: \answerNA{} 
    \item[] Justification: This work does not use crowdsourcing or human subjects.

\item {\bf Institutional review board (IRB) approvals or equivalent for research with human subjects}
    \item[] Question: Does the paper describe potential risks incurred by study participants, whether such risks were disclosed to the subjects, and whether Institutional Review Board (IRB) approvals (or an equivalent approval/review based on the requirements of your country or institution) were obtained?
    \item[] Answer: \answerNA{} 
    \item[] Justification: This work does not use crowdsourcing or human subjects.

\item {\bf Declaration of LLM usage}
    \item[] Question: Does the paper describe the usage of LLMs if it is an important, original, or non-standard component of the core methods in this research? Note that if the LLM is used only for writing, editing, or formatting purposes and does \emph{not} impact the core methodology, scientific rigor, or originality of the research, declaration is not required.
    \item[] Answer: \answerNA{} 
    \item[] Justification: We used LLM solely for editing assistance.

\end{enumerate}

\end{document}